\pdfoutput=1
\documentclass[11pt]{article}

\usepackage[final]{acl}

\usepackage{times}
\usepackage{latexsym}
\usepackage{graphicx}
\usepackage{multirow}
\usepackage{makecell}
\usepackage{subfigure}
\usepackage{algorithm}
\usepackage{algpseudocode}
\usepackage{booktabs} 
\usepackage{amssymb}
\usepackage{amsmath}
\usepackage{bbm}
\usepackage{xspace}
\usepackage{pifont}
\usepackage{colortbl}
\definecolor{shadecolor}{rgb}{0.92,0.92,0.92}
\usepackage{framed}

\usepackage[T1]{fontenc}
\usepackage[utf8]{inputenc}

\usepackage{microtype}

\usepackage{inconsolata}

\usepackage{graphicx}

\title{DaMoC: Efficiently Selecting the Optimal Large Language Model for Fine-tuning Domain Tasks Based on Data and Model Compression}

\author{
 \textbf{Wei Huang\thanks{These authors contributed equally to this work}},
 \textbf{Huang Wei$^*$},
 \textbf{Yinggui Wang}\thanks{Corresponding author (wyinggui@gmail.com).},
\\
 Ant Group, China
\\
   \{hw378176,lingxin.hw\}@antgroup.com, wyinggui@gmail.com
}

\begin{document}
\maketitle
\begin{abstract}
Large language models (LLMs) excel in general tasks but struggle with domain-specific ones, requiring fine-tuning with specific data. With many open-source LLMs available, selecting the best model for fine-tuning downstream tasks is challenging, primarily focusing on how to quickly identify the optimal LLM. We introduce a \textbf{Da}ta and \textbf{Mo}del \textbf{C}ompression Framework (DaMoC) that addresses this challenge by: 1) Data Level: A systematic categorization of data filtering methodologies for LLMs is first established, classifying them into three distinct paradigms: (1) distribution-aware methods, (2) quality-aware methods, and (3) hybrid approaches considering both dimensions. Further, we enhance the density of key tokens in the text achieving token compression. Subsequently, we use an LLM to iterative rewrite the text to optimize its expression. 2) Model Level: We use layer similarity scores to assess each layer's importance and remove those with lower importance. Then, we introduce a sparse merging paradigm to preserve as much of the original model's capability as possible. Extensive experiments on four datasets, medical Q\&A, financial Q\&A, general Q\&A, and reading comprehension, show that we can select the optimal LLM while saving approximately 20-fold in training time.

\end{abstract}

\section{Introduction}
In recent years, LLMs have rapidly developed, showcasing strong general capabilities after being pre-trained on large-scale datasets, and can handle diverse tasks through few-shot prediction or n-context learning~\cite{glm2024chatglm,huang2024prodigy}. However, in specific fields such as law and medicine, the performance of LLMs is not entirely satisfactory~\cite{yang2024zhongjing}. To improve the model's performance in these areas, users typically fine-tune the model using domain-specific data. Over the past few years, numerous institutions have released various open-source LLMs, such as Llama3~\cite{dubey2024llama}, Qwen2.5~\cite{yang2024qwen2}, etc. This brings a challenge: \textit{when fine-tuning LLMs with domain data, how should we choose the optimal open-source LLM?} Although existing research has focused on the issue of hyperparameter selection in fine-tuning LLMs~\cite{liu2024large}, no studies have yet addressed the problem of selecting an optimal LLM for fine-tuning. \textbf{The core issue we need to address is how to design an algorithm that enables rapid training and ensures relative performance stability among different models.}

\begin{figure*}[!t]
\centering
  \includegraphics[width=0.80\linewidth]{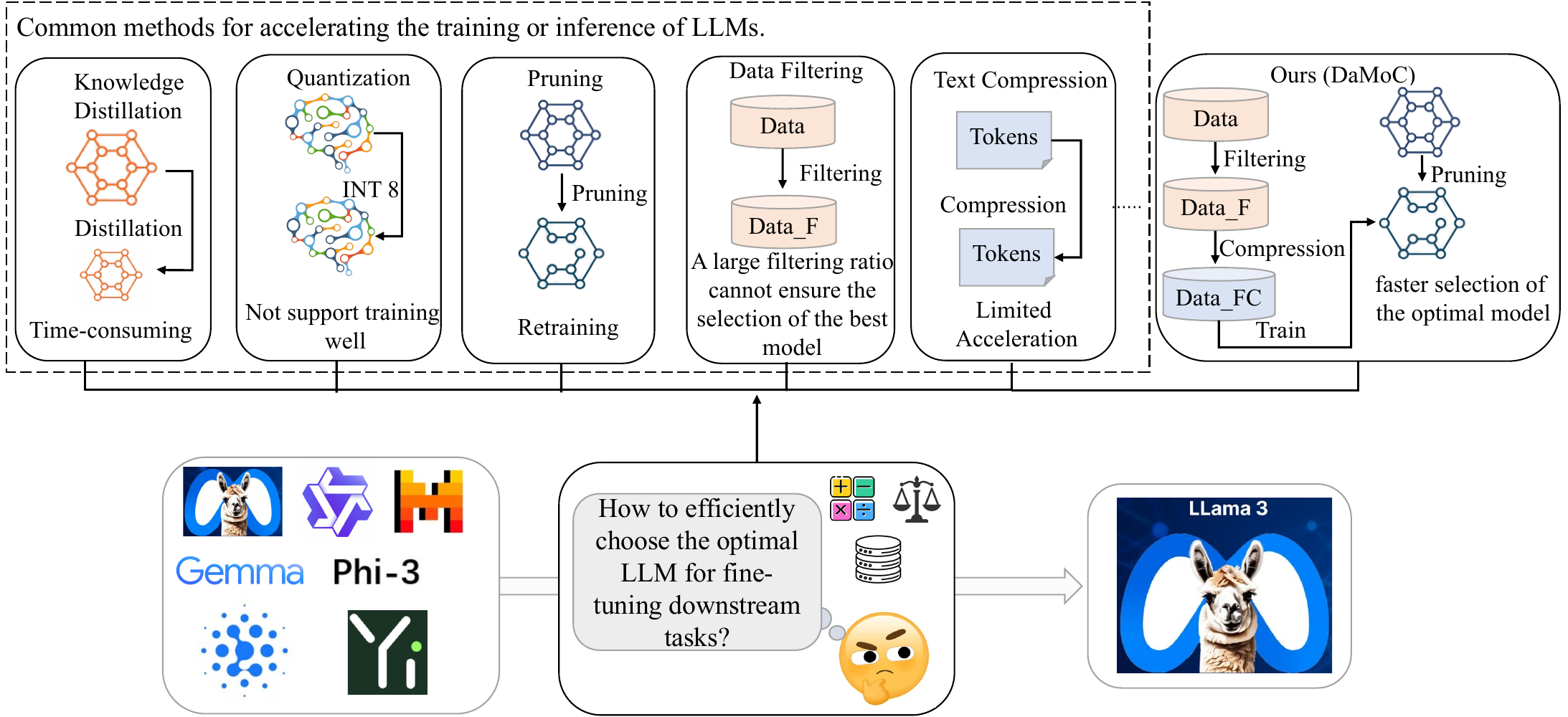}
  \caption {Illustration optimal model selection problem, the desired objectives, and methods for accelerating training.}
  \label{fig1}
  \vspace{-0.4cm}
\end{figure*}

Currently, common methods for accelerating the training or inference of LLMs include knowledge distillation~\cite{xu2024survey}, quantization~\cite{wan2023efficient}, pruning~\cite{zhu2023survey}, data filtering~\cite{chenalpagasus}, and token compression~\cite{jiang2023llmlingua}. Knowledge distillation transfers the knowledge from a teacher model to a smaller one, but the process is time-consuming. Quantization compresses the model by converting floating-point numbers to integers or fixed-point numbers. Most existing research focuses on the post-training phase, while quantization methods in the fine-tuning phase primarily focus on parameter-efficient fine-tuning, such as PEQA~\cite{kim2024memory} and Qlora~\cite{dettmers2024qlora}. Therefore, these methods are not very friendly for full fine-tuning and require hardware support. Pruning accelerates models by removing unimportant parameters, requiring a large amount of pre-training data for retraining, and there is limited exploration in scenarios where fine-tuning on downstream tasks. Data filtering selects important or similarly distributed data from the original dataset, but our experiments found that a high filtering ratio can affect the relative stability among models (see Section 5.1). Token compression speeds up processes by shortening the input text length, mainly used in the inference phase, and its application in fine-tuning still requires further exploration. A summary can be seen in Figure~\ref{fig1}.

To quickly and accurately select the best open-source LLMs for fine-tuning, we propose DaMoC. DaMoC addresses the model selection problem from both data and model perspectives. On the data level, we categorize LLM data filtering methods into three types: (1) distribution-aware methods, (2) quality-aware methods, and (3) hybrid approaches considering both dimensions. We have established a dedicated data compression framework based on these categories to evaluate the relationship between data filtering methods and optimal model selection. Through our research, we found that distribution-aware methods perform better than hybrid approaches considering both dimensions, which in turn outperform quality-aware methods(see Section 5.1 for details). Additionally, we found that data filtering methods struggle to select the optimal model when the filtering ratio is too high (>90\%). We also employed perplexity and Question-related perplexity to enhance the density of critical information in the input text. Subsequently, we used an LLM to iterative 
rewrite the text to optimize its expression achieving a compression rate of approximately 50\%. 

At the model level, we focused on model pruning to balance time efficiency and implementation generality. We adopted a layer-wise pruning approach. We began by collecting input and output activations for each layer using a small calibration dataset and calculated their cosine similarity, deeming layers with higher similarity as less important. To preserve model accuracy and parameter distribution, we introduce a
sparse merging paradigm to merge the parameters of pruned layers with those of the remaining ones.
%, which is obtained by subtracting the pre-trained model's parameters from the fine-tuned model's parameters. 
Our method successfully prunes approximately 25\% of the layers while identifying the optimal LLM.

Extensive experiments on different LLMs and various datasets show that by jointly applying data filtering, token compression, and model pruning, we can select the optimal LLM while reducing training time by approximately 20-fold. \textbf{Further analysis revealed that the DaMoC can also identify the optimal fine-tuning method from full fine-tuning and LoRA fine-tuning, providing valuable insights for future research.}
% Of course, using any one of these three optimization methods individually can also achieve optimal model selection.

% Our primary contributions are as follows:

% 1) DaMoC Framework for Optimal LLM Selection: To address the current lack of research on selecting optimal LLM during the fine-tuning, we introduce the DaMoC framework.

% 2) Explore Comprehensive Compression Strategies: DaMoC explores both data compression and model compression techniques to accelerate training on downstream tasks while ensuring the stability of performance trends.

% 3) Empirical Validation: Experimental results across eight models and four datasets demonstrate that DaMoC can identify the optimal LLM while increasing training speed by over 20 times.

\begin{table*}[!t]
\centering
\scalebox{0.58}
{\begin{tabular}{lccccc|ccccc}
\toprule
\multirow{2}{*}{Base Model} &
\multicolumn{5}{c}{PubMedQA} &
\multicolumn{5}{c}{BillSum} \\
\cmidrule(lr){2-6} \cmidrule(lr){7-11}
&Full &LoRA &Zero-shot & One-shot & Two-shot & Full &LoRA & Zero-shot & One-shot & Two-shot 
\\
\hline
Llama3.1-8B &0.5978(2)&0.6083(2) &0.4901(8)&0.5125(8)&0.5274(8) &0.7017(3)&\cellcolor{red!40}0.6958(1) &0.5689(6)&0.5943(5)&0.6002(5)   \\
Qwen2.5-7B  &0.5973(3)&\cellcolor{red!40}0.6090(1) &0.5233(5)&0.5289(5)&0.5300(7) &\cellcolor{red!40}0.7060(1)&0.6901(4) &0.5673(7)&0.5921(6)&0.5933(7)  \\
Gemma2-9B &0.5898(8)&0.6076(4) &0.5202(6)&0.5287(7)&\cellcolor{red!40}0.5358(1) &0.7016(4)&0.6893(5) &0.5509(8)&0.5782(8)&0.5891(8)   \\
GLM4-9B   &0.5914(5)&0.6046(6) &0.5253(4)&0.5318(3)&0.5341(4) &0.7021(2)&0.6948(3) &0.5798(3)&0.6012(3)&\cellcolor{red!40}0.6161(1) \\
Ministral-8B &0.5954(4)&0.6078(3) &\cellcolor{red!40}0.5296(1)&0.5344(2)&0.5353(2) &0.6979(5)& 0.6952(2)&0.5698(5)&0.5835(7)&0.5990(6)  \\
Phi-3-small-7B &\cellcolor{red!40}0.6063(1)&0.6038(8) &0.5283(2)&\cellcolor{red!40}0.5349(1) &0.5350(3) &0.6881(8)&0.6866(8) &\cellcolor{red!40}0.5947(1)&\cellcolor{red!40}0.6124(1)&0.6120(2) \\
Yi1.5-9B	& 0.5911(7)&0.6065(5) &0.5171(7)&0.5288(5)&0.5319(5) &0.6957(6)&0.6871(7) &0.5801(2)&0.6033(2)&0.6057(3) \\
Internlm2.5-7B & 0.5914(5)&0.6043(7) &0.5261(3)&0.5302(4)&0.5303(6) &0.6936(7)&0.6874(6) &0.5772(4)&0.5985(4)&0.6007(4) \\
\bottomrule
\end{tabular}}
\caption{\label{tabel1}The impact of few-shot on selecting the most suitable LLM. Full represented Full fine-tuning. LoRA represented LoRA fine-tuning. Full and LoRA serve as baselines trained on all models. The optimal model is considered to have been selected if the results of few-shot are consistent with these baselines. Numerical values in parentheses indicate the ranking of the current model's accuracy among all evaluated models, and the optimal model is denoted in red.}
\vspace{-0.3cm}
\end{table*}

\section{Motivation}
Inspired by in-context learning~\cite{dong2024survey}, we explore whether it is possible to choose the most suitable LLM for domain fine-tuning through a few-shot approach without additional training. Our experiments involve selected examples from the training data, structured as follows: $instruct + example_1 + ... + example_N + question + answer$. where N represents the number of included examples. For our experiments, we utilize two domain-specific datasets and eight LLMs. The selected datasets include PubMedQA~\cite{jin2019pubmedqa} for medical tasks and BillSum~\cite{kornilova2019billsum} for financial tasks. The chosen models encompass: 
% the most prominent open-source LLMs currently available in the field:
Llama3.1-8B~\cite{dubey2024llama}, Qwen2.5-7B~\cite{yang2024qwen2}, Gemma2-9B~\cite{team2024gemma}, GLM4-9B~\cite{glm2024chatglm}, Ministral-8B~\cite{jiang2023mistral}, Phi-3-small-7B~\cite{abdin2024phi}, Yi1.5-9B~\cite{young2024yi}, and Internlm2.5-7B~\cite{cai2024internlm2}. Through few-shot prompting, we could evaluate the performance of these models under example guidance.

It was observed from Table ~\ref{tabel1} that employing the few-shot approach to select the optimal model can result in significant errors. 
% For instance, on the PubMed dataset, the best model for the Full fine-tuning was Phi-3-small-7B, while the best model for the Zero-shot was Ministral-8B.
We guess that the primary reason for this phenomenon is that certain models may not be inherently robust for specific tasks, yet they possess strong learning capabilities. Consequently, when fine-tuned with domain-specific data, the accuracy of these models improves substantially. For example, Llama3.1-8B ranked eighth among all models when using the Zero-shot, but rose to second after fine-tuning. This raises a pertinent question: 

\textit{When fine-tuning LLMs with domain-specific data, the challenge lies in how to efficiently identify the LLMs with the highest accuracy.}

\section{Efficient LLMs Select Framework}
To quickly and accurately select the most suitable LLM for fine-tuning downstream tasks, we propose DaMoC to address this challenge. DaMoC encompasses three strategies: 1) data filtering methods, 2) token compression methods, and 3) model pruning methods. In the following sections, we will provide a detailed description.

\subsection{Data Filtering Methods in DaMoC}

% \begin{table}[!t]
% \centering
% \scalebox{0.65}{
% \begin{tabular}{lcccc}
% \toprule
% Methods & Distribution-aware & Quality-aware & Both & Runtime \\
% \midrule
% DQ & \checkmark & - & - & fast \\
% GraphCut & \checkmark & - & - & fast \\
% KCenterGreedy & \checkmark & - & - & fast \\
% Random & \checkmark & - & - & fast \\
% AlphaGasus & - & \checkmark & - & middle \\
% LMA & - & \checkmark & - & fast \\
% Superfiltering & - & \checkmark & - & middle \\
% LESS & - & \checkmark & - & slow \\
% MoDS & - & - & \checkmark & fast \\
% Cherry & - & - & \checkmark & slow \\
% Deita & - & - & \checkmark & middle \\
% CaR & - & - & \checkmark & middle \\
% \bottomrule
% \end{tabular}}
% \caption{\label{table2}The categories and runtime of different data filtering methods.}
% \end{table}
Data filtering in LLMs can reduce the quantity of data. Therefore, we explore the relationship between data filtering methods and the selection of optimal pre-trained models.
% Data filtering methodologies not only enable the selection of appropriate minimal datasets to ensure fine-tuning precision but also serve to reduce computational overhead. 
With the proliferation of various data filtering techniques for LLMs, we are compelled to ask: \textit{which data filtering methods are suitable for our scenario?} 

To systematically investigate the relationship between data filtering strategies and optimal model selection, we have developed a comprehensive framework comprising twelve distinct data filtering approaches. These methodologies have been categorized into three distinct paradigms: (1) distribution-aware methods, (2) quality-aware methods, and (3) hybrid approaches incorporating both dimensions. The rationale behind this classification is to investigate which is more crucial for selecting the optimal model: data distribution or data quality. 
% The statistics for different methods can be seen in Table ~\ref{table2}.

(1) Distribution-aware methods: \ding{172} \textbf{DQ}~\cite{zhou2023dataset} first divides the entire dataset into a set of non-overlapping bins recursively based on the submodular gains~\cite{iyer2021submodular}. Then, a small portion of data samples is uniformly sampled from all bins. \ding{173} \textbf{GraphCut}~\cite{iyer2021submodular} aims to maximize submodular gains. it expects to maximize the diversity between the sample and the selected set while minimizes the distance between the sample and the remained set. \ding{174} \textbf{KCenterGreedy(KCG)}~\cite{sener2018active} iteratively selects the point in the dataset that is farthest from the currently chosen centers as the new cluster center. \ding{175} \textbf{Random} algorithm randomly selects k sample points from the training data without replacement.

(2) Quality-aware methods: \ding{172} \textbf{AlphaGasus(AG\\)} utilizes the LLM to select high-quality data~\cite{chen2023alpagasus}. \ding{173} \textbf{LMA}~\cite{zhaolong}:selects samples with longer label from the training data. \ding{174} \textbf{Superfiltering(SF)}~\cite{li-etal-2024-superfiltering} employs GPT-2 to compute the instruction-following difficulty (IFD) score~\cite{li2024quantity} and assess the quality of the data. \ding{175} \textbf{LESS}~\cite{xialess} has constructed a reusable gradient data repository, which then selects data based on the similarity to a predefined set of examples.

(3) hybrid approaches incorporating both dimensions: \ding{172} \textbf{MoDS}~\cite{du2023mods} introduced a quality assessment model to score the data. Subsequently, a subset of the data is selected using the KCenterGreedy algorithm. \ding{173} \textbf{Cherry}~\cite{li2024quantity} Cherry employs K-Means clustering to group the data embeddings. Subsequently, n instances are sampled from each cluster for model training. The trained model is used to compute the IFD scores of the data. \ding{174} \textbf{Deita}~\cite{liumakes} first utilizes a quality scoring model to compute the quality and complexity scores for each sample. Subsequently, the k most similar data points to the current sample are identified, and these k data points are then filtered based on their scores. \ding{175} \textbf{CaR}~\cite{ge-etal-2024-clustering} uses the IQS model to evaluate the quality scores of instructions, followed by the classification of instructions through the application of the clustering methods.

The experimental results indicate that, for the tasks under study, distribution-aware methods outperform hybrid approaches incorporating both dimensions, which in turn outperform quality-aware methods. Additionally, it is noted that when the sampling rate is set to 5\%, data filtering methods tend to make a greater number of erroneous selections.
% however, at sampling rates of 10\% and 20\%, the optimal model can be correctly identified. 
The experiments also highlight that specific data filtering methods can maintain a consistent order between full fine-tuning and LoRA fine-tuning, which provides valuable insights for scenarios involving the selection of different fine-tuning strategies. For more detailed experimental analyses, please refer to Sections 5.

\subsection{Token Compression Method in DaMoC}
The previous work on token compression has primarily focused on the inference phase, with the objects of compression being instructions, reference documents, and questions, as exemplified by LLMLingua~\cite{jiang2023llmlingua}. In contrast, our primary goal is to target the compression of questions and responses during the training phase. we use $x = (x^{que}, x^{ans})$ to represent a prompt, including the instruction and the question $x^{que}$ and the answer $x^{ans}$. The objective of a prompt compression system can be formulated as:

\begin{equation}
\begin{aligned}
    \min_{\widetilde{x},x} D_{\phi}\left( Embedding(x),Embedding(\widetilde{x}) \right) 
 \end{aligned}
\end{equation}

where $\widetilde{x}$ represents the compressed prompt. $D_{\phi}$ denotes the distance function, such as Bert Score~\cite{zhang2019bertscore}. Embedding refers to the embedding model, such as Bert~\cite{devlin2018bert}.

We incorporate the iterative compression mechanism and budget controller following LLMLingua and directly calculate token perplexities to compress $x^{que}$, resulting in $\widetilde{x^{que}}$. To speed up the processing, we use the Baichuan2-7B-Chat-4bits~\cite{yang2023baichuan} model. We employ Question-related perplexity to assess the importance of each token within the $x^{ans}$. Then, tokens in response are pruned based on their importance scores, resulting in $\widetilde{x^{ans}}$. This method highlights the significance of tokens that are closely associated with the question. The formula is as follows:
\begin{equation}
\begin{aligned}
 s_i = perplexity(x_{i}^{ans}| x^{que}, x_{<i}^{ans})
 \end{aligned}
\end{equation} 
\vspace{-0.06cm}
Through the aforementioned process, we can obtain the compressed prompt $\widetilde{x} = (\widetilde{x^{que}}, \widetilde{x^{ans}})$

However, relying solely on the aforementioned token compression strategy 
% can result in text that suffers from poor readability and incoherent sentences. Sometimes, it 
may result in the loss of important information, which can cause misunderstandings during the model training. This may lead to instability in the relative performance between models, causing the selection of the suboptimal LLM. To address the above problem, for every $\widetilde{x}$ we calculate the BERTScore using Equation (1). If the score is below the threshold, we employ Baichuan2-13B-Chat for iterative text rewriting. The specific steps are as follows: 

\ding{172} A prompt was constructed in the format $\widetilde{x}^{*} = ([\widetilde{x^{que}}^{*}, \widetilde{x^{ans}}^{*}], 
[x^{que}, x^{ans}],
\widetilde{x^{doc}}^{*},
)$ where $\widetilde{x^{que}}^{*}$, and $\widetilde{x^{ans}}^{*}$ represent the question and answer that need to be rewritten. $\widetilde{x^{doc}}$ represents examples randomly selected from the compressed prompts in $\widetilde{x}$ that meet the requirements. The detailed prompt can be found in the Appendix~\ref{appdixa}. \ding{173} LLM generates a response based on the prompt. \ding{174} Once the response is generated, we calculate the BERTScore using Equation 1. If the score is above the threshold, the rewriting is considered successful. If the score is below the threshold, we update the prompt format as $\widetilde{x}^{*} = ([\widetilde{x^{que}}^{*}, \widetilde{x^{ans}}^{*}], 
[x^{que}, x^{ans}], x^{res},
\widetilde{x^{doc}}^{*},
)$. where $x^{res}$ includes the response from the current iteration and the similarity score as negative examples for the next prompt. \ding{175} Steps 2 and 3 are repeated.

It is noteworthy that token compression is based on the data filtering framework, meaning that data filtering needs to be performed first. According to the experiments, approximately 92\% of the compressed texts can directly meet the BERTScore threshold when relying solely on perplexity. Only about 8\% of the compressed texts require rewriting, and only around 0.3\% require a second round of rewriting. We use VLLM for inference, so the entire rewriting process takes very little time.

\subsection{Model Pruning Method in DaMoC}
\begin{figure}[t]
\centering
  \includegraphics[width=0.9\linewidth]{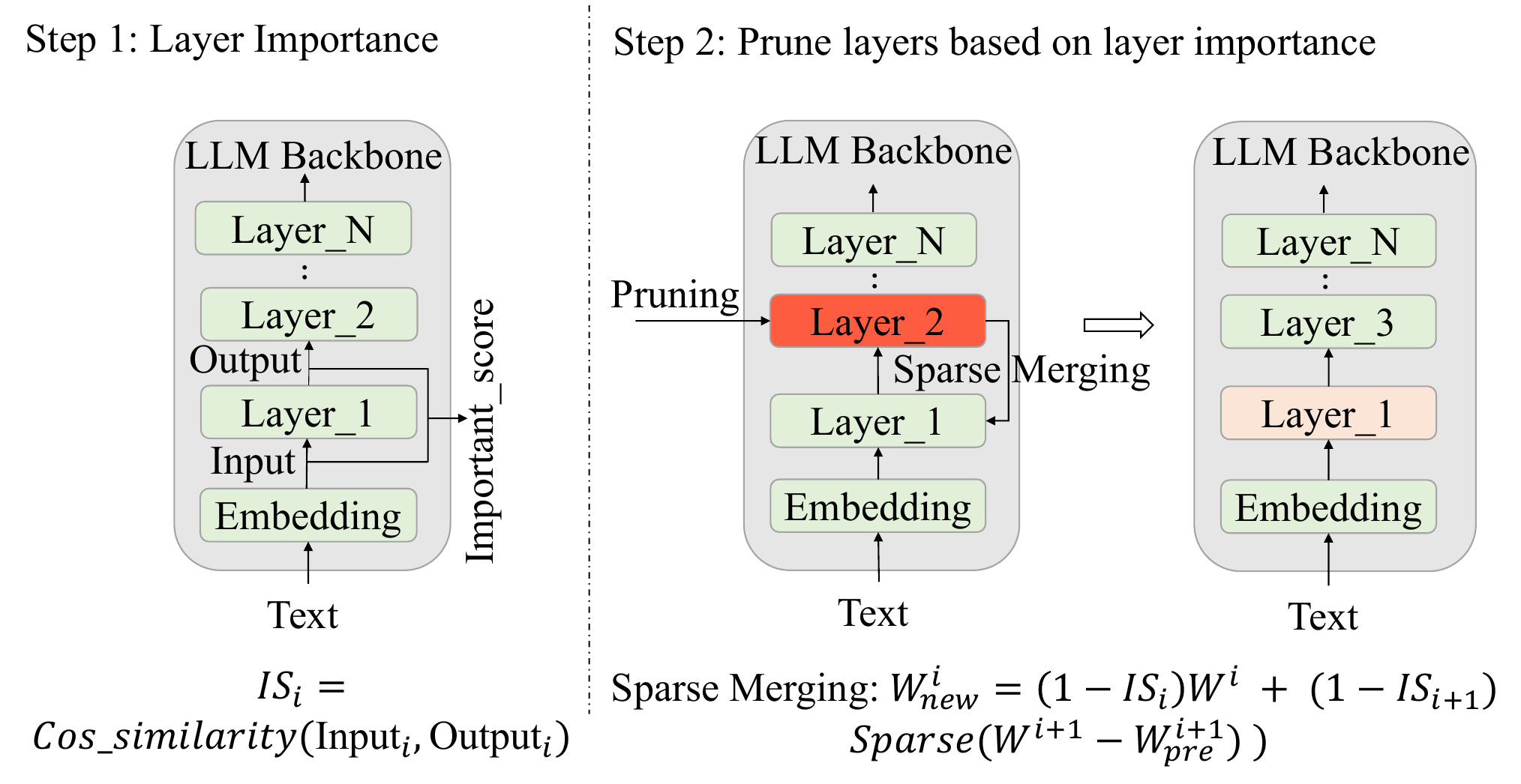}
  \caption {Flowchart of the layer pruning algorithm.}
  \label{fig2}
  \vspace{-0.3cm}
\end{figure}

In the preceding sections, we explored methods at the data level. In this section, we will shift our focus to the model level, investigating model pruning techniques. Our experiments revealed that when using ZeRO-3~\cite{rajbhandari2020zero}, the frequent inter-machine communication negated the speed benefits of pruning Multi-Head Attention (MHA), MLP, and Hidden Size(HZ) (see Appendix~\ref{appdixb} for details). Consequently, the subsequent discussion will be focused on layer-wise pruning. The flow of the algorithm can be seen in Figure ~\ref{fig2}.

% Inspired by the work of ~\cite{muralidharan2024compact}. 
In the first step, we measure the importance of model layers. Initially, a subset of calibration data is selected from the training dataset to serve as the input to the model. During the forward propagation phase, layer-wise input and output activations are collected, and the cosine similarity metric between corresponding activation pairs is computed. A high cosine similarity indicates that this layer's input and output have remained unchanged, meaning that the parameters of this layer do not cause performance fluctuations, thereby indicating lower architectural importance. We pruned the layers with a similarity score greater than the threshold. The formula is as follows:

\begin{equation}
    IS_i = \sum_{j=0}^{M} \  \frac{ X_{i,j}^{T}X_{i+1,j}}{|| X_{i,j}||_{2}|| X_{i+1,j}||_{2}}
\end{equation}
where $X_i$ refers to the input to layer $i$, and $M$ denotes the $t^{th}$ row of $X_i$. IS is Layer Importance.

Direct layer pruning induces distributional shifts in model parameters, which simultaneously degrade model accuracy and disrupt inter-model relational consistency. This makes it challenging to select the optimal model. To address these challenges, we introduce a sparse merging paradigm. Firstly, we subtract the pre-trained parameters from the parameters of the pruned layer. For example, we subtract the Base model's parameters from those of the Chat model. The resulting matrix is referred to as a task vector. Subsequently, we use the magnitude of the parameters as a sparsification metric to sparsify the task vector for pruning. Finally, we merge these pruned task vectors with the preceding unpruned layers. The merging coefficient is determined by the layer importance scores calculated using Equation (3). The purpose of sparsification is to ensure that during merging, the parameters of the pruned layers do not excessively overshadow those of the unpruned layers, which could lead to a decrease in accuracy, while still preserving the capabilities of the pruned layers. The sparse merging operation is formalized as:
\begin{equation}
\begin{split}
W_{new}^i= (1- IS_{i}) * W^i+ ( 1 - IS_{i+1}) * \\Sparse(W^{i+1} - W_{pre}^{i+1}) 
\end{split}
\end{equation}

\begin{equation}
Sparse(W_i)=\left\{
\begin{aligned}
W_i & , |W_i|>\tau \\
0 & , |W_i| \leq \tau \\
\end{aligned}
\right.
\end{equation}

\renewcommand{\arraystretch}{0.90}
\begin{table*}[!th]
\centering
\scalebox{0.50}{
\begin{tabular}{cccccccc|cccc|cccc}
\toprule
\multirow{2}{*}{Sample} &
\multirow{2}{*}{FT} &
\multirow{2}{*}{Models} &
\multirow{2}{*}{GS} &
\multicolumn{4}{c}{Distribution-aware methods} &
\multicolumn{4}{c}{Quality-aware methods} &
\multicolumn{4}{c}{Both} \\
\cmidrule(lr){5-8} \cmidrule(lr){9-12} \cmidrule(lr){13-16} ratio
& & & & DQ & GraphCut & KCG & Random & AG & LMA & SF & LESS & MoDS & Cherry & Deita & CaR \\
\hline
\multirow{16}{*}{20\%} 
&
\multirow{8}{*}{Full} 
& Llama3.1 & 0.7017(3)	&0.6791(2)&	0.6696(4)&	0.6882(2)&	0.6827(3)&	0.6683(4)	&\cellcolor{red!40}0.6636(1)	&0.6710(7)&	0.6755(5)&	\cellcolor{red!40}0.6801(1)	&0.6855(4)	&0.6713(7)&	\cellcolor{red!40}0.6793(1)\\
&& Qwen2.5 & \cellcolor{red!40}0.7060(1)	&\cellcolor{red!40}0.6829(1)&	\cellcolor{red!40}0.6727(1)&	\cellcolor{red!40}0.6902(1)&	\cellcolor{red!40}0.6869(1)	&\cellcolor{red!40}0.6706(1)	&0.6583(4)	&0.6720(4)	&\cellcolor{red!40}0.6800(1)	&0.6697(6)&	0.6900(2)&	0.6755(3)&	0.6722(3)\\
&& Gemma2 & 0.7016(4)&	0.6760(4)&	0.6671(8)&	0.6845(3)&	0.6811(5)&	0.6639(6)&	0.6568(5)	&0.6719(5)&	0.6699(8)	&0.6712(4)&	\cellcolor{red!40}0.6913(1)&	0.6697(8)	&0.6711(4)\\
&& GLM4 & 0.7021(2)	&0.6759(5)	&0.6675(7)	&0.6780(6)	&0.6772(7)	&0.6697(3)	&0.6606(2)&	0.6716(6)&	0.6748(7)&	0.6691(7)&	0.6851(5)&	0.6722(5)	&0.6677(7)\\
&& Ministral& 0.6979(5)	&0.6667(8)	&0.6584(5)	&0.6787(5)	&0.6817(4)	&0.6576(8)	&0.6598(3)	&\cellcolor{red!40}0.6778(1)&	0.6756(3)	&0.6740(3)	&0.6805(8)&	0.6759(2)	&0.6711(4)\\
&& Phi-3-small& 0.6881(8)	&0.6761(3)&	0.6697(3)&	0.6770(8)&	0.6771(8)	&0.6705(2)	&0.6551(6)	&0.6651(8)	&0.6751(6)	&0.6699(5)	&0.6858(3)	&\cellcolor{red!40}0.6768(1)	&0.6703(6)\\
&& Yi1.5      & 0.6957(6)	&0.6728(6)&	0.6676(6)	&0.6815(4)&	0.6850(2)&	0.6657(5)&	0.6534(7)	&0.6767(2)	&0.6778(2)&	0.6752(2)	&0.6812(7)&	0.6729(4)&	0.6749(2)\\
&& Internlm2.5& 0.6936(7)	&0.6716(7)	&0.6721(2)	&0.6772(7)&	0.6806(6)	&0.6577(7)	&0.6469(8)	&0.6724(3)	&0.6756(4)	&0.6679(8)	&0.6817(6)&	0.6714(6)	&0.6601(8)\\

\cmidrule(lr){2-16}
&
\multirow{8}{*}{LoRA} 
& Llama3.1 & \cellcolor{red!40}0.6958(1)&	0.6737(2)&	\cellcolor{red!40}0.6694(1)&\cellcolor{red!40}	0.6805(1)	&\cellcolor{red!40}0.6817(1)&	0.6645(5)&	0.6542(4)	&0.6716(3)&	0.6834(2)&	0.6783(2)&	0.6852(3)&	0.6668(5)	&0.6718(2)\\
&& Qwen2.5 & 0.6901(4)&	0.6651(7)	&0.6663(5)	&0.6760(4)	&0.6683(6)&	0.6648(4)&	0.6494(5)&	0.6634(6)	&0.6760(5)&	0.6675(7)&	0.6774(6)	&0.6661(6)	&0.6660(7)\\
&& Gemma2 & 0.6893(5)	&0.6730(5)	&0.6677(4)&	0.6738(6)&	0.6789(4)&	\cellcolor{red!40}0.6730(1)&	0.6600(3)&\cellcolor{red!40}0.6793(1)	&0.6782(4)	&0.6724(3)	&0.6835(4)	&0.6770(3)	&0.6668(5)\\
&& GLM4 & 0.6948(3)	&0.6732(3)&	0.6683(3)	&0.6779(3)&	0.6793(3)&	0.6619(7)&	0.6638(2)	&0.6720(2)	&0.6795(3)	&0.6721(4)&\cellcolor{red!40}	0.6864(1)	&\cellcolor{red!40}0.6819(1)	&0.6710(3)\\
&& Ministral & 0.6952(2)	&\cellcolor{red!40}0.6758(1)&	0.6690(2)	&0.6798(2)&	0.6805(2)&	0.6635(6)	&\cellcolor{red!40}0.6656(1)	&0.6824(5)&	\cellcolor{red!40}0.6860(1)&\cellcolor{red!40}	0.6790(1)	&0.6859(2)&	0.6775(2)&	\cellcolor{red!40}0.6760(1)\\
&& Phi-3-small & 0.6866(8)	&0.6650(8)	&0.6580(6)	&0.6644(8)	&0.6624(8)&	0.6650(3)&	0.6410(7)&	0.6533(8)	&0.6638(8)	&0.6623(8)	&0.6712(8)	&0.6594(8)	&0.6630(8)\\
&& Yi1.5 & 0.6871(7)&	0.6731(4)&	0.6527(8)	&0.6750(5)	&0.6732(5)	&0.6584(8)	&0.6484(6)	&0.6702(4)	&0.6700(6)	&0.6691(6)	&0.6813(5)&	0.6706(4)&	0.6689(4)\\
&& Internlm2.5 &0.6874(6)&	0.6691(6)	&0.6574(7)&	0.6674(7)	&0.6675(7)&	0.6656(2)&	0.6356(8)	&0.6601(7)	&0.6680(7)	&0.6702(5)&	0.6756(7)	&0.6635(7)&	0.6664(6)\\
\hline
\hline
\multirow{16}{*}{10\%} 
&
\multirow{8}{*}{Full} 
& Llama3.1 & 0.7017(3)	&0.6716(3)	&0.6688(3)	&0.6824(2)&	0.6764(3)	&\cellcolor{red!40}0.6657(1)&	\cellcolor{red!40}0.6511(1)&	0.6656(3)&	0.6717(3)&	0.6666(2)	&0.6729(8)	&0.6679(3)	&0.6605(4)\\
&& Qwen2.5 &\cellcolor{red!40} 0.7060(1)	&\cellcolor{red!40}0.6797(1)	&\cellcolor{red!40}0.6729(1)	&\cellcolor{red!40}0.6884(1)	&\cellcolor{red!40}0.6789(1)	&0.6616(5)	&0.6501(3)	&0.6664(2)&	0.6721(2)&	\cellcolor{red!40}0.6728(1)&	0.6810(2)&	0.6654(4)&	\cellcolor{red!40}0.6680(1)\\
&& Gemma2 & 0.7016(4)&	0.6699(5)&	0.6675(5)&	0.6776(3)&	0.6732(6)&0.6603(6)	&0.6462(6)&\cellcolor{red!40}	0.6679(1)	&0.6661(6)&	0.6637(4)	&0.6785(4)&\cellcolor{red!40}	0.6706(1)&	0.6656(2)\\
&& GLM4 & 0.7021(2)	&0.6718(2)	&0.6717(2)&	0.6694(8)&	0.6746(4)&	0.6640(3)&	0.6506(2)&	0.6637(5)	&\cellcolor{red!40}0.6725(1)	&0.6632(5)	&0.6782(5)&	0.6630(6)&	0.6599(6)\\
&& Ministral & 0.6979(5)&	0.6658(6)&	0.6585(8)&	0.6757(5)&	0.6737(5)&	0.6645(2)&	0.6475(4)	&0.6546(8)	&0.6645(7)&	0.6583(7)&	0.6790(3)	&0.6654(4)&	0.6598(7)\\
&& Phi-3-small & 0.6881(8)	&0.6657(7)	&0.6654(6)	&0.6728(6)	&0.6693(8)	&0.6594(7)	&0.6467(5)&	0.6602(6)&	0.6698(4)&	0.6582(8)&	0.6767(6)	&0.6590(7)	&0.6552(8)\\
&& Yi1.5 & 0.6957(6)&	0.6701(4)&	0.6644(7)&	0.6766(4)&0.6779(2)	&	0.6637(4)&	0.6420(7)&	0.6647(4)&	0.6674(5)&	0.6644(3)&	\cellcolor{red!40}0.6821(1)&	0.6699(2)&	0.6600(5)\\
&& Internlm2.5 & 0.6936(7)	&0.6597(8)	&0.6680(4)	&0.6701(7)	&0.6730(7)	&0.6566(8)	&0.6341(8)	&0.6578(7)	&0.6684(8)	&0.6594(6)	&0.6762(7)	&0.6582(8)&	0.6621(3)\\

\cmidrule(lr){2-16}
&
\multirow{8}{*}{LoRA} 
& Llama3.1 & \cellcolor{red!40}0.6958(1)	&\cellcolor{red!40}0.6994(1)&\cellcolor{red!40}	0.6674(1)&\cellcolor{red!40}	0.6809(1)&\cellcolor{red!40}	0.6749(1)&	0.6587(5)&	0.6476(3)&	0.6629(4)&	\cellcolor{red!40}0.6772(1)&	0.6661(4)&	0.6768(3)	&0.6626(6)&	0.6548(6)\\
&& Qwen2.5 & 0.6901(4)&	0.6607(7)&	0.6555(7)&	0.6678(6)&	0.6636(7)	&0.6634(2)&	0.6405(5)&	0.6605(5)&	0.6652(5)	&0.6475(7)&	0.6685(7)&	0.6659(5)&	0.6510(7)\\
&& Gemma2 & 0.6893(5)&	0.6665(4)&	0.6601(5)&	0.6787(2)&	0.6705(4)&	0.6614(4)&	0.6522(2)	&0.6636(3)	&0.6700(4)	&0.6682(3)&	0.6779(2)&	0.6661(4)	&0.6652(3)\\
&& GLM4 & 0.6948(3)&	0.6719(3)	&0.6652(3)	&0.6733(4)	&0.6703(5)&	0.6620(3)&	\cellcolor{red!40}0.6626(1)&	\cellcolor{red!40}0.6766(1)&	0.6763(2)	&0.6735(2)	&\cellcolor{red!40}0.6789(1)	&\cellcolor{red!40}0.6751(1)&\cellcolor{red!40}	0.6721(1)\\
&& Ministral & 0.6952(2)&	0.6733(2)	&0.6666(2)	&0.6780(3)	&0.6722(3)	&\cellcolor{red!40}0.6657(1)&	0.6469(4)	&0.6696(2)	&0.6733(3)&	\cellcolor{red!40}0.6760(1)&	0.6746(4)&	0.6740(2)&	0.6667(2)\\
&& Phi-3-small & 0.6866(8)&	0.6548(8)	&0.6398(8)&	0.6506(8)	&0.6634(8)	&0.6527(6)	&0.6264(8)&	0.6524(8)&	0.6512(8)&	0.6432(8)&	0.6591(8)	&0.6490(8)	&0.6451(8)\\
&& Yi1.5 & 0.6871(7)&	0.6657(5)	&0.6625(4)&	0.6655(7)	&0.6734(2)&	0.6508(8)&	0.6392(6)	&0.6590(6)&	0.6554(7)	&0.6638(5)	&0.6692(6)	&0.6695(3)&	0.6607(4)\\
&& Internlm2.5 & 0.6874(6)	&0.6634(6)	&0.6575(6)	&0.6682(5)	&0.6676(6)	&0.6525(7)&	0.6315(7)	&0.6572(7)&	0.6602(6)	&0.6562(6)	&0.6708(5)&	0.6527(7)	&0.6553(5)\\
\hline
\hline
\multirow{16}{*}{5\%} 
&
\multirow{8}{*}{Full} 
& Llama3.1 & 0.7017(3)	&0.6663(2)	&0.6479(7)	&0.6647(3)	&0.6671(3)	&\cellcolor{red!40}0.6637(1)	&0.6367(2)&	0.6487(5)&	0.6633(2)	&\cellcolor{red!40}0.6737(1)	&0.6703(3)	&0.6593(3)&	\cellcolor{red!40}0.6581(1)\\
&& Qwen2.5 & \cellcolor{red!40}0.7060(1)&	\cellcolor{red!40}0.6675(1)&	0.6523(4)	&\cellcolor{red!40}0.6672(1)&	0.6676(2)&	0.6589(2)&	\cellcolor{red!40}0.6420(1)&	0.6468(7)&\cellcolor{red!40}	0.6682(1)&	0.6670(2)&	0.6708(2)&	\cellcolor{red!40}0.6650(1)&	0.6535(5)\\
&& Gemma2 & 0.7016(4)&	0.6623(3)&	0.6556(2)&	0.6536(8)&	0.6570(7)&	0.6582(3)&	0.6306(7)&	0.6504(4)&	0.6617(3)&	0.6571(8)&\cellcolor{red!40}	0.6712(1)&	0.6569(5)&	0.6530(7)\\
&& GLM4 & 0.7021(2)	&0.6606(4)&	0.6495(6)&	0.6564(7)&	0.6644(4)&	0.6563(6)&	0.6313(6)&	0.6510(3)&	0.6578(6)&	0.6629(4)&	0.6686(4)&	0.6521(7)&	0.6536(4)\\
&& Ministral & 0.6979(5)&	0.6572(5)&	0.6450(8)&	0.6642(4)&	0.6640(5)&	0.6550(8)&	0.6348(3)&	0.6422(8)&	0.6538(8)&	0.6574(5)&	0.6640(8)	&0.6497(8)&	0.6534(6)\\
&& Phi-3-small & 0.6881(8)&	0.6569(6)&	0.6540(3)&	0.6603(6)&	0.6541(8)&	0.6569(5)&	0.6314(5)&	0.6469(6)&	0.6599(4)&	0.6558(6)&	0.6678(5)&	0.6551(6)&	0.6427(8)\\
&& Yi1.5 & 0.6957(6)&	0.6568(7)&	0.6516(5)&	0.6622(5)&\cellcolor{red!40}	0.6681(1)&	0.6575(4)&	0.6345(4)&	\cellcolor{red!40}0.6598(1)&	0.6588(5)&	0.6659(3)&	0.6665(7)&	0.6597(2)&	0.6558(2)\\
&& Internlm2.5 & 0.6936(7)&	0.6603(8)&\cellcolor{red!40}	0.6558(1)&	0.6660(2)&	0.6610(6)&	0.6558(7)&	0.6277(8)&	0.6512(2)&	0.6569(7)&	0.6537(7)&	0.6669(6)&	0.6592(4)&	0.6552(3)\\
\cmidrule(lr){2-16}
&
\multirow{8}{*}{LoRA}
& Llama3.1 & \cellcolor{red!40}0.6958(1)&	0.6660(2)&	0.6496(4)&	0.6588(4)&	0.6576(4)&	0.6571(3)&	0.6260(6)&	0.6498(5)&	0.6613(4)&	0.6479(5)&	0.6701(4)&	0.6545(6)&	0.6478(5)\\
&& Qwen2.5 & 0.6901(4)&	0.6416(7)&	0.6387(7)&	0.6457(7)&	0.6488(8)&	0.6521(5)&	0.6280(5)&	0.6487(6)&	0.6476(7)&	0.6402(7)&	0.6551(7)&	0.6548(5)&	0.6382(7)\\
&& Gemma2 & 0.6893(5)&	0.6623(4)&	0.6578(3)&	0.6677(3)&	0.6695(2)&	0.6525(4)&	0.6370(2)&	0.6538(3)&	0.6660(2)&	0.6533(3)&	0.6705(3)&	0.6624(3)&	0.6537(3)\\
&& GLM4 & 0.6948(3)&	0.6658(3)&\cellcolor{red!40}	0.6623(1)&	0.6697(2)&	0.6602(3)&	0.6659(2)&	\cellcolor{red!40}0.6553(1)&\cellcolor{red!40}	0.6664(1)&	0.6642(3)&	\cellcolor{red!40}0.6632(1)&	\cellcolor{red!40}0.6789(1)	&\cellcolor{red!40}0.6771(1)	&\cellcolor{red!40}0.6689(1)\\
&& Ministral & 0.6952(2)&	\cellcolor{red!40}0.6715(1)&	0.6598(2)&	\cellcolor{red!40}0.6739(1)	&\cellcolor{red!40}0.6722(1)&\cellcolor{red!40}	0.6670(1)&	0.6296(4)&	0.6516(4)&\cellcolor{red!40}	0.6686(1)&	0.6550(2)&	0.6707(2)	&0.6680(2)&	0.6591(2)\\
&& Phi-3-small & 0.6866(8)&	0.6371(8)&	0.6333(8)&	0.6417(8)	&0.6336(7)	&0.6469(7)&	0.6053(8)&	0.6434(7)	&0.6338(8)&	0.6328(8)&	0.6511(8)&	0.6389(8)&	0.6322(8)\\
&& Yi1.5 & 0.6871(7)&	0.6529(6)&	0.6490(5)&	0.6561(5)	&0.6574(5)&	0.6520(6)&	0.6336(3)	&0.6602(2)&	0.6610(5)	&0.6449(6)&	0.6695(5)&	0.6592(4)	&0.6488(4)\\
&& Internlm2.5 & 0.6874(6)	&0.6556(5)	&0.6459(6)	&0.6552(6)	&0.6534(6)	&0.6458(8)	&0.6146(7)&	0.6379(8)&	0.6487(6)	&0.6485(4)&	0.6594(6)&	0.6451(7)	&0.6423(6)\\
\bottomrule
\end{tabular}}
\caption{The results of data filtering in the BillSum dataset. FT donated Fine-tuning. Both represented hybrid approaches incorporating both dimensions. Due to space constraints in the main paper, the results for the remaining three datasets are provided in Appendix~\ref{appdixD}.}
\label{table3}
\vspace{-0.3cm}
\end{table*}

\section{Experimental Settings}
\paragraph{Datasets and Setting.} 
We evaluated DaMoC on eight widely adopted Large Language Models: Llama3.1-8B, Qwen2.5-7B, Gemma2-9B, GLM4-9B, Ministral-8B, Phi-3-small-7B, Yi1.5-9B, and Internlm2.5-7B and four key tasks: medical Q\&A, financial Q\&A, general Q\&A, and
reading comprehension. The datasets used in this study include PubMedQA~\cite{jin2019pubmedqa}, BillSum~\cite{kornilova2019billsum}, Alpaca~\cite{alpaca}, and SQuAD~\cite{rajpurkar-etal-2016-squad}. For the Alpaca data, we use the Base model for fine-tuning. The remaining data is fine-tuned using the Instruct model. Detailed descriptions of the datasets and the evaluation methods are provided in Appendix~\ref{appdixg}. The setup of our experimental can be found in Appendix ~\ref{es}.

% Further, we utilized the Spearman rank coefficient~\cite{Arsov2019AMO} to measure the similarity of ranking between data selection methods and baselines. The higher the similarity, the better the data preserves the trend of the model.

\paragraph{Compared Methods.}
1) Grid Search(GS): Fine-tuning is conducted for eight models and four datasets. Our latter two optimization strategies consist of token compression and layer pruning. To demonstrate the advantages of our proposed token compression and layer pruning, we are integrating the existing methods into our data filtering framework for comparison.
2) LLMLingua~\cite{jiang2023llmlingua}: Perplexity is used for iterative token compression, while a budget controller is employed to maintain semantic integrity. 3) LongLLMLingua~\cite{jiang2023longllmlingua}: A problem-aware coarse-to-fine compression method is proposed to enhance the density of critical information in prompts. 4) Laco~\cite{yang2024laco}: The rear model layers are collapsed into a prior layer, enabling a rapid reduction in model size.

\renewcommand{\arraystretch}{0.90}
\begin{table*}[!th]
\centering
\scalebox{0.60}{
\begin{tabular}{ccccccc|ccc}
\toprule
\multirow{2}{*}{Dataset} &
\multirow{2}{*}{FT} &
\multirow{2}{*}{Models} &
\multirow{2}{*}{GS} &
\multicolumn{3}{c}{GraphCut} &
\multicolumn{3}{c}{Random} \\
\cmidrule(lr){5-7} \cmidrule(lr){8-10}
& & & &DaMoC &LLMLingua+Laco  &LongLLMLingua+Laco & DaMoC &LLMLingua+Laco  &LongLLMLingua+Laco \\
\hline
\multirow{16}{*}{PubMedQA}
&
\multirow{8}{*}{Full}
&Llama3.1 &\cellcolor{red!40}0.5978(1) &\cellcolor{red!40}0.5918(1)&\cellcolor{red!40}0.5671(1)&\cellcolor{red!40}0.5681(1) &\cellcolor{red!40}0.5923(1)&0.5588(2)&\cellcolor{red!40}0.5611(1)\\
&&Qwen2.5 &0.5973(2)  &0.5894(2)&0.5654(2)&0.5651(2) &0.5870(2)&\cellcolor{red!40}0.5601(1)&0.5600(2)\\
&&Gemma2 &0.5898(6) &0.5798(6)&0.5612(5)&0.5622(4) &0.5844(4)&0.5536(6)&0.5533(6) \\
&&GLM4 &0.5914(3)    &0.5842(3)&0.5644(3)&0.5640(3) &0.5847(3)&0.5547(4)&0.5557(3)\\
&&Yi1.5 &0.5911(5)  &0.5811(5)&0.5589(6)&0.5601(6) &0.5835(5)&0.5569(3)&0.5549(5)\\
&&Internlm2.5 &0.5914(3) &0.5838(4)&0.5629(4)&0.5618(5) &0.5808(6)&0.5547(4)&0.5555(4)\\
\cmidrule(lr){2-10}
&
\multirow{8}{*}{LoRA}
&Llama3.1 &0.6083(2)  &0.6003(3)&0.5828(2)&\cellcolor{red!40}0.5874(1)&0.6033(1)&0.5810(2)&0.5835(2)\\
&&Qwen2.5 &\cellcolor{red!40}0.6090(1)  &\cellcolor{red!40}0.6047(1)&\cellcolor{red!40}0.5862(1)&0.5869(2)&\cellcolor{red!40}0.6033(1)&\cellcolor{red!40}0.5839(1)&\cellcolor{red!40}0.5851(1)\\
&&Gemma2 &0.6076(3) &0.6010(2)&0.5714(5)&0.5748(6)&0.5982(3)&0.5717(6)&0.5730(6) \\
&&GLM4 &0.6046(5) &0.5969(5)&0.5785(3)&0.5783(4)&0.5978(4)&0.5766(4)&0.5789(3)\\
&&Yi1.5 &0.6065(4) &0.5978(4)&0.5766(4)&0.5772(5)&0.5964(5)&0.5734(5)&0.5736(5)\\
&&Internlm2.5 &0.6043(6) &0.5939(6)&0.5810(3)&0.5835(3)&0.5912(6)&0.5795(3)&0.5767(4)\\
\hline
\hline
\multirow{16}{*}{Billsum}
&
\multirow{8}{*}{Full}
&Llama3.1 &0.7017(3) &0.6433(4)&\cellcolor{red!40}0.6257(1)&\cellcolor{red!40}0.6305(1)&0.6453(2)&0.6212(2)&\cellcolor{red!40}0.6218(1)\\
&&Qwen2.5 &\cellcolor{red!40}0.7060(1)  &\cellcolor{red!40}0.6522(1)&0.6139(5)&0.6161(5)&\cellcolor{red!40}0.6518(1)&0.6183(4)&0.6194(5)\\
&&Gemma2 &0.7016(4) &0.6439(3)&0.6202(2)&0.6234(2)&0.6437(4)&\cellcolor{red!40}0.6217(1)&0.6218(1) \\
&&GLM4 &0.7021(2)     &0.6448(2)&0.6183(3)&0.6183(4)&0.6425(3)&0.6175(5)&0.6199(4)\\
&&Yi1.5 &0.6957(5)  &0.6415(5)&0.6155(4)&0.6159(6)&0.6406(5)&0.6166(6)&0.6186(6)\\
&&Internlm2.5 &0.6936(6) &0.6394(6)&0.6187(6)&0.6228(3)&0.6389(6)&0.6200(3)&0.6212(3)\\
\cmidrule(lr){2-10}
&
\multirow{8}{*}{LoRA}
&Llama3.1 &\cellcolor{red!40}0.6958(1)  &\cellcolor{red!40}0.6419(1)&0.6235(5)&0.6252(4)&0.6405(3)&0.6225(5)&0.6252(2)\\
&&Qwen2.5 &0.6901(3)   &0.6410(2)&0.6284(2)&\cellcolor{red!40}0.6300(1)&0.6406(2)&0.6250(2)&0.6252(2)\\
&&Gemma2 &0.6893(4)  &0.6393(5)&0.6230(6)&0.6222(6)&0.6391(4)&0.6201(6)&0.6144(6) \\
&&GLM4 &0.6948(2)    &0.6405(3)&\cellcolor{red!40}0.6288(1)&0.6288(2)&\cellcolor{red!40}0.6419(1)&0.6244(3)&0.6147(5)\\
&&Yi1.5 &0.6871(6) &0.6399(4)&0.6283(3)&0.6275(3)&0.6373(6)&\cellcolor{red!40}0.6278(1)&\cellcolor{red!40}0.6258(1)\\
&&Internlm2.5 &0.6874(5) &0.6385(6)&0.6240(4)&0.6250(5)&0.6376(5)&0.6231(4)&0.6184(4)\\
\hline
\hline
\multirow{16}{*}{SQuAD}
&
\multirow{8}{*}{Full}
&Llama3.1 &\cellcolor{red!40}0.6586(1) &\cellcolor{red!40}0.6444(1)&0.5914(4)&0.5960(3)&\cellcolor{red!40}0.6296(1)&0.5814(5)&0.5846(5)\\
&&Qwen2.5 &0.6581(3)  &0.6319(2)&0.5941(3)&0.5946(4)&0.6288(2)&0.5903(3)&\cellcolor{red!40}0.5932(1)\\
&&Gemma2 &0.6428(5)  &0.6164(5)&\cellcolor{red!40}0.5977(1)&\cellcolor{red!40}0.5988(1)&0.6255(4)&0.5905(2)&0.5899(4) \\
&&GLM4 &0.5927(6)    &0.5335(6)&0.5089(6)&0.5172(6)&0.5778(6)&0.5210(6)&0.5225(6)\\
&&Yi1.5 &0.6550(4) &0.6180(4)&0.5826(5)&0.5746(5)&0.6203(5)&0.5835(4)&0.5907(3)\\
&&Internlm2.5 &0.6586(1) &0.6287(3)&0.5944(2)&0.5961(2)&0.6269(3)&\cellcolor{red!40}0.5917(1)&0.5919(2)\\
\cmidrule(lr){2-10}
&
\multirow{8}{*}{LoRA}
&Llama3.1 &0.6624(4) &0.6538(4)&\cellcolor{red!40}0.6257(1)&0.6323(2)&0.6490(5)&0.6233(2)&\cellcolor{red!40}0.6274(1)\\
&&Qwen2.5 &0.6633(3)  &0.6557(3)&0.6236(3)&\cellcolor{red!40}0.6245(1)&0.6530(4)&\cellcolor{red!40}0.6236(1)&0.6258(2)\\
&&Gemma2 &0.6669(2)  &\cellcolor{red!40}0.6574(1)&0.6249(2)&0.6303(3)&\cellcolor{red!40}0.6660(1)&0.6195(4)&0.6210(3) \\
&&GLM4 &\cellcolor{red!40}0.6685(1)    &0.6569(2)&0.6233(4)&0.6285(5)&0.6574(3)&0.6193(5)&0.6210(3)\\
&&Yi1.5 &0.6573(5) &0.6519(5)&0.6189(6)&0.6240(6)&0.6588(2)&0.6199(3)&0.6214(4)\\
&&Internlm2.5 &0.6565(6) &0.6488(6)&0.6202(5)&0.6300(4)&0.6474(6)&0.6190(6)&0.6211(5)\\
\bottomrule
\end{tabular}}
\caption{The comparative results of DaMoC with different token compression and layer pruning methods. We selected the two data filtering methods that performed best in the previous experiments, GraphCut and Random, as the foundation for our subsequent experiments. Our layer pruning algorithm requires $W_{pre}$. Since our fine-tuning of Alpaca is based on the base model, and neither Phi-3-small nor Ministral-8B has released their base models, obtaining 
$W_{pre}$ is not feasible. Therefore, we only report the results on six models and three datasets.
% Suppose users are fine-tuning base model or are using a model without a released base version. In that case, it is recommended that they use data filtering combined with token compression, as evidenced by the results in Section 6.
}
\label{table4}
\vspace{-0.3cm}
\end{table*}

\section{Results}
% In this section, we provide the experimental results to verify the effectiveness and robustness of DaMoC.

\subsection{Results of the Data Filtering Framework} 
In Table ~\ref{table3}, we present the results of the Data Filtering Framework on BillSum. In this Framework, we evaluated various sampling rates, specifically 20\%, 10\%, and 5\%. The results indicate that Distribution-aware methods consistently select the optimal pre-trained model. In contrast, the Distribution-aware method exhibits a higher number of incorrect selections. Therefore, for the task of selecting the optimal model, Distribution-aware methods outperform hybrid approaches that incorporate both dimensions, which in turn outperform Quality-aware methods. On the other hand, it was also observed that when the sampling rate was 5\%, data filtering methods exhibited a higher frequency of errors. In contrast, at sampling rates of 10\% and 20\%, the optimal model could still be selected. We guess that this is primarily because the model does not learn sufficiently with too little data, leading to fluctuations in stability. 
% Furthermore, it is noted that no single model achieves the highest accuracy across all datasets. For instance, the model with the highest accuracy on the PubMedQA dataset is Phi-3-small, whereas Qwen2.5 exhibits the highest accuracy on the BillSum dataset. This observation underscores that a model's superior performance on one dataset does not necessarily translate to optimal performance on another.
% it has been observed that certain data filtering methods effectively maintain the order between Full fine-tuning and LoRA fine-tuning. This suggests that when faced with uncertainty in choosing between different fine-tuning methods, our framework can swiftly and accurately identify the method that yields the highest accuracy.

\subsection{Results of the DaMoC}

% We selected the two data filtering methods that performed best in the previous experiments, GraphCut and Random, as the foundation for our subsequent experiments. 
Table ~\ref{table4} presents the comparison results of DaMoC and different methods. The comparison methods are also based on our data filtering framework. Since DaMoC also includes token compression and layer pruning, we combined the comparative methods while ensuring that the levels of token compression and pruning were kept consistent with DaMoC. The experimental results demonstrate that DaMoC exhibits superior performance, generally ensuring the selection of the optimal pre-trained model. Additionally, the method based on GraphCut slightly outperforms the one based on Random. The Random-based method occasionally selected suboptimal models. The comparison methods showed terrible effects for the Billsum and SQuAD. Further analysis revealed that even when the DaMoC has a significant impact on accuracy, it can still select the optimal model, as evidenced by the experimental results on the BillSum dataset. This demonstrates that our proposed method maintains stability across different models. When users are selecting the optimal pre-trained model for fine-tuning downstream tasks. If fine-tuning is based on a Chat model, we suggested approach comprises GraphCut, Token Compression, and Model Pruning; conversely, if fine-tuning is conducted on a Base model, the recommended strategy involves GraphCut and Token Compression.

\subsection{Ablation Study and Runtime}

\renewcommand{\arraystretch}{0.90}
\begin{table*}[!th]
\centering
\scalebox{0.55}{
\begin{tabular}{cccc|c|c|cc}
\toprule
\multirow{2}{*}{Data} &
\multirow{2}{*}{FT} &
\multirow{2}{*}{Models} &
\multirow{2}{*}{GS} &
\multirow{2}{*}{only MP} &
\multirow{2}{*}{only TC} &
\multicolumn{2}{c}{GraphCut} \\
\cmidrule(lr){7-8}
& & & & & & DaMoC w/o TC &DaMoC w/o MP \\
\hline
\multirow{16}{*}{PubMedQA}
&
\multirow{8}{*}{Full}
&Llama3.1 &\cellcolor{blue!40}0.5978(2) &\cellcolor{blue!40}0.5913(1)&0.5959(3)&\cellcolor{blue!40}0.5914(1)&0.5989(2)\\
&&Qwen2.5 &0.5973(3)  &0.5892(2)&0.5977(2)&0.5883(2)&0.5974(3)\\
&&Gemma2 &0.5898(8)  &0.5800(6)&0.5913(7)&0.5792(6)&0.5913(6)\\
&&GLM4 &0.5914(5)    &0.5836(5)&0.5862(8)&0.5835(4)&0.5859(8)\\
&&Ministral &0.5954(4) &-&0.5921(5)&-&0.5923(4)\\
&&Phi-3-small &\cellcolor{red!40}0.6063(1) &-&\cellcolor{red!40}0.6011(1)&-&\cellcolor{red!40}0.6013(1)\\
&&Yi1.5 &0.5911(7) &0.5839(4)&0.5927(4)&0.5812(5)&0.5916(5)\\
&&Internlm2.5 &0.5914(5) &0.5845(3)&0.5914(6)&0.5843(3)&0.5906(7)\\

\cmidrule(lr){2-8}
&
\multirow{8}{*}{LoRA}
&Llama3.1 &0.6083(2) &0.6001(3)&0.6047(3)&0.6000(3)&0.6050(2)\\
&&Qwen2.5 &\cellcolor{red!40}0.6090(1)  &\cellcolor{red!40}0.6038(1)&\cellcolor{red!40}0.6054(1)&\cellcolor{red!40}0.6041(1)&\cellcolor{red!40}0.6055(1)\\
&&Gemma2 &0.6076(4)  &0.6003(2)&0.6053(2)&0.6005(2)&0.6045(3)\\
&&GLM4 &0.6046(6)    &0.5959(5)&0.6022(6)&0.5969(5)&0.6033(5)\\
&&Ministral &0.6078(3) &-&0.6036(4)&-&0.6040(4)\\
&&Phi-3-small &0.6040(8) &-&0.6004(8)&-&0.6002(8)\\
&&Yi1.5 &0.6065(5) &0.5968(4)&0.6018(7)&0.5977(4) &0.6027(7)\\
&&Internlm2.5 &0.6043(7) &0.5934(6)&0.6035(5)&0.5933(6)&0.6031(6)\\
\bottomrule
\end{tabular}}
\caption{Results of the ablation experiments in PubMedQA. Here, MP denotes Model Prune, TC signifies Token Compression and DaMoC w/o TC represents Data Filtering combined with Model Pruning. Since model pruning requires 
$W_{pre}$, and some models, such as Phi-3-small, do not disclose 
$W_{pre}$. To differentiate, when comparing the performance of MP and DaMoC w/o TC, we use blue to indicate the optimal model. If the optimal model overlaps with red, we still use red. The results for the remaining datasets can be found in Appendix ~\ref{appdixE}.}
\label{table22}
\vspace{-10pt}
\end{table*}

\paragraph{Ablation Study.}
Our ablation study includes four aspects: 1) using only Token Compression (only TC), 2) using only Model Pruning (only MP), 3) Data Filtering + Model Pruning (DaMoC w/o TC), and 4) Data Filtering + Token Compression (DaMoC w/o MP). As mentioned in the previous section, the most suitable data filtering method for model selection is GraphCut. Therefore, our ablation experiments are based on this method. 
% We presented the results of the remaining three datasets in Appendix~\ref{appdixE}. 
The experimental results in Table~\ref{table22} demonstrate that the Token Compression method and Model Pruning technique in DaMoC are both effective in selecting the most suitable pre-trained model for fine-tuning downstream tasks. This indicates that both methods are capable of maintaining the relative stability of accuracy across different models.

\paragraph{Cost Savings for Train.} To calculate the time savings of DaMoC during the training phase, a subset of 3000 samples was selected from the training data for testing purposes. During the testing phase, training was conducted for only one epoch. The training time for the 3000 samples was approximately 15.28 minutes, whereas DaMoC required about 0.72 minutes. This resulted in a time saving of approximately 20-fold. Additionally, we present the token compression ratios for different data sets, as shown in Table~\ref{table5}, and the number of pruned layers for different models, as shown in Table ~\ref{table6}.

\renewcommand{\arraystretch}{0.90}
\begin{table}[!t]
\centering
\scalebox{0.70}{
\begin{tabular}{ccccc}
\toprule
\multirow{2}{*}{Methods} &
\multicolumn{4}{c}{GraphCut} \\
\cmidrule(lr){2-5}
& PubMedQA &BillSum &Alpaca &SQuAD \\
\hline
Graphcut&40.66\%&62.31\%&48.32\%&25.37\% \\
Random &41.72\%	&62.16\%	&49.61\%	&24.51\% \\
\bottomrule
\end{tabular}}
\caption{Token compression ratios across different datasets.}
\label{table5}
\vspace{-0.3cm}
\end{table}

\renewcommand{\arraystretch}{0.90}
\begin{table}[!t]
\centering
\scalebox{0.60}{
\begin{tabular}{ccccccc}
\toprule
\multirow{2}{*}{Methods} &
\multicolumn{6}{c}{Models} \\
\cmidrule(lr){2-7}
& Llama3.1	&Qwen2.5	&Gemma2	&GLM4 & Yi1.5 & Internlm2.5 \\
\hline
quantity &8/32 &7/28 &10/42 &10/40 &11/48 &7/28 \\

\bottomrule
\end{tabular}}
\caption{The number of pruned layers in different models. The number before the slash indicates the number of pruned layers, while the after one represents the original number of layers in the model.}
\label{table6}
\vspace{-0.3cm}
\end{table}

\renewcommand{\arraystretch}{0.90}
\begin{table}[!th]
\centering
\scalebox{0.55}{
\begin{tabular}{cccc|cc}
\toprule
\multirow{2}{*}{Dataset} &
\multirow{2}{*}{Models} &
\multicolumn{2}{c}{GS} &
\multicolumn{2}{c}{DaMoC} \\
\cmidrule(lr){3-4} \cmidrule(lr){5-6}
&  &Full &LoRA  &Full &LoRA \\
\hline
\multirow{8}{*}{PubMedQA}
&Llama3.1 &0.5978(2)&\cellcolor{red!40}0.6083(1)&0.5918(2)&\cellcolor{red!40}0.6003(1)\\
&Qwen2.5 &0.5973(2)&\cellcolor{red!40}0.6090(1)&0.5894(2)&\cellcolor{red!40}0.6047(1) \\
&Gemma2 &0.5898(2)&\cellcolor{red!40}0.6076(1)&0.5798(2)&\cellcolor{red!40}0.6010(1)\\
&GLM4 &0.5914(2) &\cellcolor{red!40}0.6046(1)&0.5842(2)&\cellcolor{red!40}0.5969(1)\\
&Yi1.5 &0.5911(2)&\cellcolor{red!40}0.6065(1) &0.5811(2)&\cellcolor{red!40}0.5978(1)\\
&Internlm2.5 &0.5914(2)&\cellcolor{red!40}0.6043(1)&0.5838(2) &\cellcolor{red!40}0.5939(1)\\

\hline
\hline
\multirow{8}{*}{Billsum}
&Llama3.1 &\cellcolor{red!40}0.7017(1)&0.6958(2)&\cellcolor{red!40}0.6433(1)&0.6419(2)\\
&Qwen2.5 &\cellcolor{red!40}0.7060(1)&0.6901(2)&\cellcolor{red!40}0.6522(1)&0.6410(2)\\
&Gemma2 &\cellcolor{red!40}0.7016(1)&0.6893(2)&\cellcolor{red!40}0.6439(1)&0.6393(2)\\
&GLM4 &\cellcolor{red!40}0.7021(1)&0.6948(2)&\cellcolor{red!40}0.6448(1)&0.6405(2)\\
&Yi1.5 &\cellcolor{red!40}0.6957(1)&0.6871(2)&\cellcolor{red!40}0.6415(1)&0.6399(2)\\
&Internlm2.5 &\cellcolor{red!40}0.6936(1)&0.6874(2)&\cellcolor{red!40}0.6394(1)&0.6385(2)\\

\hline
\hline
\multirow{8}{*}{SQuAD}
&Llama3.1 &0.6586(2)&\cellcolor{red!40}0.6624(1)&0.6444(2)&\cellcolor{red!40}0.6538(1)\\
&Qwen2.5 &0.6581(2)&\cellcolor{red!40}0.6633(1)&0.6319(2)&\cellcolor{red!40}0.6557(1)\\
&Gemma2 &0.6428(2)&\cellcolor{red!40}0.6669(1)&0.6164(2)&\cellcolor{red!40}0.6574(1)\\
&GLM4 &0.5927(2)&\cellcolor{red!40}0.6685(1)&0.5335(2)&\cellcolor{red!40}0.6569(1)\\
&Yi1.5 &0.6550(2)&\cellcolor{red!40}0.6573(1)&0.6180(2)&\cellcolor{red!40}0.6519(1)\\
&Internlm2.5 &\cellcolor{red!40}0.6586(1)&0.6565(2)&0.6287(2)&\cellcolor{red!40}0.6488(1)\\
\bottomrule
\end{tabular}}
\caption{The application of DaMoC in the selection of fine-tuning methods.}
\label{table7}
\vspace{-0.4cm}
\end{table}

\subsection{DaMoC for Fine-Tuning Method Selection}

Readers may naturally inquire whether the DaMoC can be used when selecting fine-tuning methods.
% (considering only task accuracy and not memory savings).
To answer this question, we conduct relevant experiments. As shown in Table ~\ref{table7}, the experiments demonstrate that when choosing between Full fine-tuning and LoRA fine-tuning, the DaMoC framework consistently identifies the optimal fine-tuning method across eight models and three datasets. These findings indicate that the DaMoC can provide valuable insights for research on selecting the most effective fine-tuning method.

\section{Related Work}
%Our methodology is primarily comprised of three components: data filtering, token compression, and model pruning. 

%Data filtering serves to identify and select data that is either crucial or advantageous for downstream training tasks. 
We classify data filtering approaches into three primary categories.
% distribution-aware methods, quality-aware methods, and hybrid methods that integrate both perspectives. 
\textbf{(\romannumeral 1)}. Distribution-aware methods prioritize data diversity. 
For instance,  GraphCut \cite{iyer2021submodular} seeks to enhance the diversity between samples and the selected dataset while minimizing the distance between samples and the remaining dataset. 
\textbf{(\romannumeral 2)} Quality-aware methods leverage quality assessment models to evaluate the quality of data.
%, followed by a filtering process. 
For example, Superfiltering(SF)~\cite{li-etal-2024-superfiltering} computes the IFD scores of data, subsequently applying filters based on these scores. 
\textbf{(\romannumeral 3)} Hybrid approaches incorporating both dimensions to refine data filtering. An illustrative example is provided by CaR~\cite{ge-etal-2024-clustering}, which initially evaluates data quality scores and subsequently applies clustering methods to achieve diversity classification of instructions. 

Token compression is predominantly utilized to curtail the textual length of prompts in models. For instance,  LongLLMLingua ~\cite{jiang2023longllmlingua} designed to address the challenges of noise and positional bias in extended contexts by enhancing the large language model's sensitivity to critical information pertinent to the query. Model pruning aims to reduce the size of models by removing parameters deemed non-essential, with a particular focus on pre-trained models. For instance, FLAP \cite{An_Zhao_Yu_Tang_Wang_2024} evaluates the variation of each input feature concerning a baseline value to estimate the consequences of removing columns of weights.  

\section{Conclusion}
We introduce the DaMoC to address the problem of selecting the optimal LLM for fine-tuning. The DaMoC addresses this challenge from three aspects: 1) Data Filtering: We investigate the impact of data filtering methods on the selection of the optimal model. 2) Token Compression: Employ perplexity to compress tokens and use iterative rewriting to enhance the expressiveness. 3) Model Pruning: We combine layer importance and sparse merging to prune the model's layers. 
The results demonstrate that DaMoC can select the optimal LLM while saving up to 20-fold the training time.

\section*{Limitations}
To address the challenge of rapidly and accurately selecting the optimal model for downstream task fine-tuning, we propose the DaMoC framework. DaMoC employs layer pruning to accelerate training; however, it requires $W_{pre}$ parameters. If the $W_{pre}$ parameters are not released by the model provider or if the user is fine-tuning based on a base model, the layer pruning algorithm within the DaMoC framework cannot be utilized. In such cases, only data filtering and token compression can be applied. Future research will focus on further exploring pruning algorithms that do not depend on $W_{pre}$ parameters.

Currently, when training large models, users face a variety of fine-tuning method choices. Although we have initially demonstrated that DaMoC can effectively select appropriate fine-tuning methods from full fine-tuning and LoRA fine-tuning, its performance with a broader range of fine-tuning methods has yet to be explored. This will be a focus of our future research.

%\multicolumn{1}{l}{}    & \multicolumn{2}{c}{Spearman rank coefficient}                              & 1.0000       & \multicolumn{1}{l}{}        & \multicolumn{1}{l}{}         & \multicolumn{1}{l}{}         & \multicolumn{1}{l|}{}                              & \multicolumn{1}{l}{}         & \multicolumn{1}{l}{}         & \multicolumn{1}{l}{}         & \multicolumn{1}{l}{}                              & \multicolumn{1}{l}{}         & \multicolumn{1}{l}{}         & \multicolumn{1}{l}{}        & \multicolumn{1}{l}{}        \\ 

\bibliography{custom}
\clearpage
\appendix
\section{The Prompt Used for Text Rewriting}
\label{appdixa}

\begin{shaded}
    You are an intelligent text optimization assistant. I have a piece of [Text] composed of questions and answers and a corresponding [Compressed Text]. The [compressed text] is obtained by removing certain content from the [text]. Your task is to appropriately optimize the [compressed text] based on the provided example and the requirements outlined below to meet the specified needs.
    
    [Text]
    
    {To be filled with the specific text}
    
    [Compressed Text]
    
    {To be filled with the specific text}
    
    [Examples]
    
    {To be filled with the specific examples}
    
    [Requirements]:
    
    The objective is to enhance the compressed text's expressiveness while ensuring logical coherence and improved comprehensibility, without expanding its content. The revised text must maintain semantic consistency with the [Text]. The length of the modified text is controlled to remain within ±10\% of the original compressed text length.
    
    Follow these procedures: 
    
    1) Thoroughly analyze the provided Text and Examples. 
    
    2) Strictly adhere to the specified requirements.
    
    3) Reference the given examples during revision.
    
    4) Execute textual modification and verify compliance with all criteria.
    
    The template for your response is as follows:
    
    Revised Compressed Text: {Only the revised text can be provided here.}
    
    Reason: {Only the reason for the revision can be provided here.}
    
    Please think step by step.
\end{shaded}

\section{The result of MHA, MLP, and HS Prune}
\label{appdixb}

\renewcommand{\arraystretch}{0.90}
\begin{table}[!th]
\centering
\scalebox{0.60}{
\begin{tabular}{ccc}
\toprule
Methods	&PubMedQA	&BillSum \\
\hline
The model before pruning. &0.53h&1.2h \\
The model after pruning. &0.48h&1.04h \\

\bottomrule
\end{tabular}}
\caption{The impact of pruning MHA, MLP, and HZ on training acceleration is examined. Our experiments were conducted based on Llama3.1. The complete training time is reported.}
\label{table8}
\end{table}

In reference to the work of~\cite{muralidharan2024compact}, the pruning of the model's Multi-Head Attention (MHA), MLP, and Hidden Size (HZ) was conducted. The hyperparameters for pruning were adopted from the original paper. The effects of training time reduction using Zero3 are reported, as illustrated in Table ~\ref{table8}. From the table, it can be observed that pruning MHA, MLP, and HZ does not result in significant acceleration of training when Zero3 is applied. Consequently, subsequent experiments were conducted based solely on layer pruning.

\section{Supplementary Experimental Setup}
\subsection{Data and evaluation methods}
\label{appdixg}
We utilize the PubMedQA datasets( 61.30k training
samples)~\cite{jin2019pubmedqa} for medical tasks, with (BERTScore + Rouge)/2~\cite{zhang-etal-2024-balancing} as the evaluation metric. For financial tasks, we employ the BillSum datasets (18.9k training
samples)~\cite{kornilova2019billsum}, with (BERTScore + Rouge)/2 as the evaluation metric. For general Q\&A, we use the Alpaca dataset(52k training
samples)~\cite{alpaca}, with MMLU(0-shot) for evaluation. Lastly, we utilize the SQuAD dataset (87.6k
training samples)~\cite{rajpurkar-etal-2016-squad} for reading comprehension,
with the F1-score as the evaluation metric. We used the base model for training for the Alpaca data, while we used the chat model for the rest of the data.

\subsection{Experimental Setting}
\label{es}
During training, we set the learning rate to 1e-5 and the batch size to 64. Each dataset was trained for 3 epochs. The AdamW optimizer was used for finetuning. We employed SWIFT~\cite{zhao2024swiftascalablelightweightinfrastructure} as the training platform and vLLM~\cite{kwon2023efficient} for inference. We follow the hyperparameter settings from the original paper for the data filtering framework. Through experimentation, it has been indicated that similarity score threshold values between 0.85 and 0.90 yield satisfactory results (see Appendix~\ref{appdixc}). In this paper, a default value of 0.85 is adopted. We set the threshold of BERTScore is 0.9. We set the sparsity rate is 20\% for sparse merging. \textit{To eliminate experimental bias, we set two distinct random seeds and presented the average accuracy results in the paper.}

\section{The results for different pruning thresholds}
\label{appdixc}

\renewcommand{\arraystretch}{0.90}
\begin{table}[!th]
\centering
\scalebox{0.60}{
\begin{tabular}{ccccc}
\toprule
Models &GS &0.80 &0.85 &0.90\\
\hline

Llama3.1 &\cellcolor{red!40}0.5978(1)  &0.5642(3)&\cellcolor{red!40}0.5918(1)&\cellcolor{red!40}0.5921(1)\\
Qwen2.5 &0.5973(2)  &\cellcolor{red!40}0.5684(1)&0.5894(2)&0.5894(2)\\
Gemma2 &0.5898(6)  &0.5575(5)&0.5798(6)&0.5814(6)\\
GLM4 &0.5914(3)    &0.5669(2)&0.5842(3)&0.5858(3)\\
Yi1.5 &0.5911(5) &0.5537(6)&0.5811(5)&0.5813(5)\\
Internlm2.5 &0.5914(3) &0.5578(4)&0.5838(4)&0.5826(4)\\
\bottomrule
\end{tabular}}
\caption{The experimental results for different similarity score thresholds. The data used in the table is from PubMedQA, and full fine-tuning was employed.}
\label{table9}
\end{table}

Ablation experiments were conducted using different similarity score thresholds, and the results are presented in Table~\ref{table9}. The findings indicate that when the similarity score threshold is set to 0.8, DaMoC fails to obtain the optimal pre-trained model. We hypothesize that this is primarily due to the excessive number of pruned layers, which significantly impacts the model's performance, leading to the selection of a suboptimal solution. Therefore, in subsequent experiments, the threshold was set to 0.85.

\section{Additional Results of the Data Filtering Framework}
\label{appdixD}
Tables~\ref{table10},~\ref{table11}, and~\ref{table12} present the experimental results of the data filtering framework on the PubMedQA, BillSum, and SQuAD datasets. From these tables, it can be observed that the GraphCut and Random methods are generally capable of selecting the optimal pre-trained model.

\renewcommand{\arraystretch}{0.90}
\begin{table*}[!th]
\centering
\scalebox{0.60}{
\begin{tabular}{cccc|c|c|cc}
\toprule
\multirow{2}{*}{Data} &
\multirow{2}{*}{FT} &
\multirow{2}{*}{Models} &
\multirow{2}{*}{GS} &
\multirow{2}{*}{only MP} &
\multirow{2}{*}{only TC} &
\multicolumn{2}{c}{GraphCut} \\
\cmidrule(lr){7-8}
& & & & & & DaMoC w/o TC &DaMoC w/o MP \\
\hline
\multirow{16}{*}{Billsum}
&
\multirow{8}{*}{Full}
&Llama3.1 &0.7017(3) &0.6848(2)&0.6551(4)&0.6838(4)&0.6558(3)\\
&&Qwen2.5 &\cellcolor{red!40}0.7060(1)  &\cellcolor{red!40}0.6879(1)&\cellcolor{red!40}0.6635(1)&\cellcolor{red!40}0.6855(1)&\cellcolor{red!40}0.6628(1)\\
&&Gemma2 &0.7016(4)  &0.6835(4)&0.6550(5)&0.6839(3)&0.6557(4)\\
&&GLM4 &0.7021(2)    &0.6848(2)&0.6563(2)&0.6843(2)&0.6560(2)\\
&&Ministral &0.6979(5) &-&0.6549(6)&-&0.6550(5)\\
&&Phi-3-small &0.6881(8) &-&0.6422(8)&-&0.6413(8)\\
&&Yi1.5 &0.6957(6) &0.6774(5)&0.6557(3)&0.6744(5)&0.6550(5)\\
&&Internlm2.5 &0.6936(7) &0.6745(6)&0.6547(7)&0.6732(6)&0.6546(7)\\

\cmidrule(lr){2-8}
&
\multirow{8}{*}{LoRA}
&Llama3.1 &\cellcolor{red!40}0.6958(1)  &\cellcolor{red!40}0.6941(1)&0.6533(1)0&\cellcolor{red!40}0.6836(1)&\cellcolor{red!40}0.6529(1)\\
&&Qwen2.5 &0.6901(4)  &0.6799(4)&0.6484(5)&0.6793(4)&0.6481(5)\\
&&Gemma2 &0.6893(5)  &0.6832(3)&0.6521(3)&0.6821(2)&0.6503(4)\\
&&GLM4 &0.6948(3)    &0.6835(2)&0.6520(4)&0.6805(3)&0.6510(3)\\
&&Ministral &0.6952(2)  &-&0.6534(2)&-&0.6515(2)\\
&&Phi-3-small &0.6866(8) &-&0.6249(8)&-&0.6222(8)\\
&&Yi1.5 &0.6871(7) &0.6757(6)&0.6471(7)&0.6744(6)&0.6470(7)\\
&&Internlm2.5 &0.6874(6) &0.6766(5)&0.6489(6)&0.6752(5)&0.6482(6)\\
\bottomrule
\end{tabular}}
\caption{Results of the ablation experiments in BillSum.}
\label{table13}
\end{table*}

\renewcommand{\arraystretch}{0.90}
\begin{table*}[!th]
\centering
\scalebox{0.60}{
\begin{tabular}{cccc|c|c|cc}
\toprule
\multirow{2}{*}{Data} &
\multirow{2}{*}{FT} &
\multirow{2}{*}{Models} &
\multirow{2}{*}{GS} &
\multirow{2}{*}{only MP} &
\multirow{2}{*}{only TC} &
\multicolumn{2}{c}{GraphCut} \\
\cmidrule(lr){7-8}
& & & & & & DaMoC w/o TC &DaMoC w/o MP \\
\hline
\multirow{16}{*}{Alpaca}
&
\multirow{8}{*}{Full}
&Llama3.1 &0.5621(6) &-&0.5634(5)&-&0.5625(5)\\
&&Qwen2.5 &\cellcolor{red!40}0.6885(1)  &-&\cellcolor{red!40}0.6884(1)&-&\cellcolor{red!40}0.6887(1)\\
&&Gemma2 &0.5699(5)   &-&0.5622(6)&-&0.5613(6)\\
&&GLM4 &0.6497(3)    &-&0.6493(3)&-&0.6499(3)\\
&&Ministral &0.4033(7) &-&0.4076(7)&-&0.4127(7)\\
&&Phi-3-small &- &-&-&-&-\\
&&Yi1.5 &0.6407(4) &-&0.6445(4)&-&0.6438(4)\\
&&Internlm2.5 &0.6685(2) &-&0.6671(2)&-&0.6677(2)\\

\cmidrule(lr){2-8}
&
\multirow{8}{*}{LoRA}
&Llama3.1 &0.6855(6) &-&0.6830(6)&-&0.6838(6)\\
&&Qwen2.5 &0.7298(2)  &-&0.7275(2)&-&0.7284(2)\\
&&Gemma2 &\cellcolor{red!40}0.7389(1)  &-&\cellcolor{red!40}0.7366(1)&-&\cellcolor{red!40}0.7367(1)\\
&&GLM4 &0.7254(3)    &-&0.7236(3)&-&0.7244(3)\\
&&Ministral &0.6213(7)  &-&0.6191(7)&-&0.6229(7)\\
&&Phi-3-small &- &-&-&-&-\\
&&Yi1.5 &0.6956(4) &-&0.6868(5)&-&0.6854(5)\\
&&Internlm2.5 &0.6875(5) &-&0.6852(4)&-&0.6833(4)\\
\bottomrule
\end{tabular}}
\caption{Results of the ablation experiments in Alpaca.}
\label{table14}
\end{table*}

% \renewcommand{\arraystretch}{0.90}
% \begin{table}[!th]
% \centering
% \scalebox{0.51}{
% \begin{tabular}{cccc|c|c|cc}
% \toprule
% \multirow{2}{*}{Data} &
% \multirow{2}{*}{FT} &
% \multirow{2}{*}{Models} &
% \multirow{2}{*}{GS} &
% \multirow{2}{*}{TC} &
% \multirow{2}{*}{MP} &
% \multicolumn{2}{c}{GraphCut} \\
% \cmidrule(lr){7-8}
% & & & & & & DaMoC w/o TC &DaMoC w/o MP \\
% \hline
% \multirow{16}{*}{Pub}
% &
% \multirow{8}{*}{Full}
% &Llama3.1 &0.6586(1)  &&&&\\
% &&Qwen2.5 &0.6581(3)  &&&&\\
% &&Gemma2 &0.6428(6)  &&&&\\
% &&GLM4 &0.5927(8)     &&&&\\
% &&Ministral &0.6191(7) &&&&\\
% &&Phi-3-small &0.6514(5) &&&&\\
% &&Yi1.5 &0.6550(4) &&&&\\
% &&Internlm2.5 &0.6586(1) &&&&\\

% \cmidrule(lr){2-8}
% &
% \multirow{8}{*}{LoRA}
% &Llama3.1 &0.6624(5) &&&&\\
% &&Qwen2.5 &0.6633(4)  &&&&\\
% &&Gemma2 &0.6669(3)  &&&&\\
% &&GLM4 &0.6685(2)    &&&&\\
% &&Ministral &0.6707(1)  &&&&\\
% &&Phi-3-small &0.6544(8) &&&&\\
% &&Yi1.5 &0.6573(6) &&&&\\
% &&Internlm2.5 &0.6565(7)  &&&&\\
% \bottomrule
% \end{tabular}}
% \caption{Results of the ablation experiments in SQuAD.}
% \label{table15}
% \end{table}

\renewcommand{\arraystretch}{0.90}
\begin{table*}[!th]
\centering
\scalebox{0.51}{
\begin{tabular}{cccccccc|cccc|cccc}
\toprule
\multirow{2}{*}{Sr} &
\multirow{2}{*}{FT} &
\multirow{2}{*}{Models} &
\multirow{2}{*}{GS} &
\multicolumn{4}{c}{Distribution-aware methods} &
\multicolumn{4}{c}{Quality-aware methods} &
\multicolumn{4}{c}{Both} \\
\cmidrule(lr){5-8} \cmidrule(lr){9-12} \cmidrule(lr){13-16} ratio
& & & & DQ & GraphCut & KCG & Random & AG & LMA & SF & LESS & MoDS & Cherry & Deita & CaR \\
\hline
\multirow{16}{*}{20\%} 
&
\multirow{8}{*}{Full} 
& Llama3.1 & 0.5978(2)& 0.5938(2) & 0.5957(2)&0.5942(3)&0.5949(2)&0.5942(2)&0.5680(3)&0.5722(3)&0.5942(3)&0.5975(2)&0.5938(2)&0.5806(3)&0.5965(2) \\
&& Qwen2.5 & 0.5973(3)& 0.5934(3) & 0.5956(3)&0.5938(4)&0.5946(3)&0.5916(3)&0.5710(2)&0.5755(2)&0.5961(2)&0.5941(3)&0.5938(2)&0.5826(2)&0.5939(3)\\
&& Gemma2 &  0.5898(8)&0.5919(5)  &0.5898(7)&0.5921(5)&0.5901(7)&0.5868(7)&0.5649(5)&0.5685(5)&0.5906(6)&0.5916(6)&0.5909(5)&0.5756(7)&0.5887(8)\\
&& GLM4 &    0.5914(5)&0.5894(7)&0.5901(6)&0.5890(6)&0.5920(5)&0.5878(6)&0.5648(6)&0.5652(8)&0.5873(8)&0.5917(5)&0.5905(6)&0.5770(6)&0.5916(5)\\
&& Ministral&0.5954(4)&0.5925(4)&0.5925(4)&0.5950(2)&0.5929(4)&0.5902(4)&0.5666(4)&0.5681(7)&0.5914(5)&0.5938(4)&0.5920(4)&0.5787(4)&0.5928(4)\\
&& Phi-3-small&\cellcolor{red!40}0.6063(1)&\cellcolor{red!40}0.6070(1)&\cellcolor{red!40}0.6058(1)&\cellcolor{red!40}0.6060(1)&\cellcolor{red!40}0.6045(1)&\cellcolor{red!40}0.6060(1)&\cellcolor{red!40}0.5796(1)&\cellcolor{red!40}0.5952(1)&\cellcolor{red!40}0.6042(1)&\cellcolor{red!40}0.6030(1)&\cellcolor{red!40}0.6037(1)&\cellcolor{red!40}0.5953(1)&\cellcolor{red!40}0.6043(1)\\
&& Yi1.5      &0.5911(7)&0.5898(6)&0.5912(5)&0.5881(8)&0.5904(6)&0.5901(5)&0.5646(7)&0.5720(4)&0.5927(4)&0.5914(7)&0.5894(7)&0.5779(5)&0.5891(7)\\
&& Internlm2.5&0.5914(5)&0.5884(8)&0.5879(8)&0.5885(7)&0.5886(8)&0.5853(8)&0.5561(8)&0.5682(6)&0.5886(7)&0.5893(8)&0.5891(8)	&0.5705(8)	&0.5893(6)\\

\cmidrule(lr){2-16}
&
\multirow{8}{*}{LoRA} 
& Llama3.1 & 0.6083(2)&	0.6053(3)&	0.6071(2)&	\cellcolor{red!40}0.6083(1)	&0.6054(3)&	0.6064(3)&	\cellcolor{red!40}0.5768(1)&	0.5912(3)&	\cellcolor{red!40}0.5996(1)&	0.6072(3)&	\cellcolor{red!40}0.6074(1)&	0.6040(3)&	0.6020(3) \\
&& Qwen2.5 & \cellcolor{red!40} 0.6090(1)	&\cellcolor{red!40}0.6068(1)&	\cellcolor{red!40}0.6083(1)&	0.6076(2)&	\cellcolor{red!40}0.6076(1)&	\cellcolor{red!40}0.6087(1)	&0.5712(2)&	\cellcolor{red!40}0.5989(1)&	0.5944(2)&	\cellcolor{red!40}0.6092(1)&	0.6070(2)&\cellcolor{red!40}	0.6062(1)&	\cellcolor{red!40}0.6045(1) \\
&& Gemma2 & 0.6076(4)&	0.6048(5)&	0.6062(4)&	0.6034(6)&	0.6042(4)&	0.6043(6)&	0.5695(4)&	0.5896(4)&	0.5898(6)&	0.6071(4)&	0.6060(4)&	0.6001(4)&	0.6001(6)\\
&& GLM4 & 0.6046(6)&	0.6052(4)&	0.6054(5)&	0.6053(4)&	0.6022(7)&	0.6058(4)&	0.5687(5)&	0.5886(5)&	0.5923(4)&	0.6049(6)&	0.6034(6)&	0.5993(5)&	0.6018(4)\\
&& Ministral & 0.6078(3)&	0.6058(2)&	0.6065(3)&	0.6071(3)&	0.6065(2)&	0.6082(2)&	0.5702(3)&	0.5945(2)&	0.5943(3)&	0.6083(2)&	0.6058(3)&	0.6057(2)&	0.6034(2)\\
&& Phi-3-small & 0.6040(8)	&0.6034(6)&	0.6032(7)&	0.6015(7)&	0.6010(8)&	0.6023(8)&	0.5656(7)&	0.5860(7)&	0.5892(7)&	0.6028(8)&	0.6023(8)&	0.5962(7)&	0.5993(8)\\
&& Yi1.5 & 0.6065(5)&	0.5993(7)&	0.6050(6)&	0.6042(5)&	0.6038(5)&	0.6050(5)&	0.5669(6)&	0.5877(6)&	0.5902(5)&	0.6055(5)&	0.6042(5)&	0.5981(6)&	0.6007(5)\\
&& Internlm2.5 & 0.6043(7)&	0.5987(8)&	0.6023(8)&	0.6010(8)&	0.6027(6)&	0.6033(7)&	0.5589(8)&	0.5851(8)&	0.5890(8)&	0.6031(7)&	0.6027(7)&	0.5951(8)&	0.5996(7)\\
\hline
\hline
\multirow{16}{*}{10\%} 
&
\multirow{8}{*}{Full} 
& Llama3.1 & 0.5978(2)	&0.5950(3)&0.5955(2)&0.5946(2)&0.5952(2)&0.5911(3)&0.5614(2)&0.5698(3)&0.5962(2)&0.5935(3)&0.5931(2)&0.5805(3)&0.5939(3) \\
&& Qwen2.5 & 0.5973(3)&0.5958(2)&0.5946(3)&0.5937(3)&0.5924(4)&0.5935(2)&0.5609(3)&0.5826(2)&0.5951(3)&0.5961(2)&0.5927(4)&0.5811(2)&0.5959(2)\\
&& Gemma2 &  0.5898(8)&0.5892(6)&0.5912(6)&0.5741(8)&0.5904(6)&0.5877(5)&0.557(6)&0.5663(6)&0.5904(6)&0.5902(4)	&0.5917(5)&0.5746(7)&0.5903(6)\\
&& GLM4 &  0.5914(5)&0.5899(5)&0.5893(7)&0.5877(7)&0.5899(7)&0.5875(6)&0.5563(7)& 0.5644(8)&0.5897(7)& 0.5892(6)&0.5882(7)&0.5751(6)&0.5901(7)\\
&& Ministral &  0.5954(4) &	0.5916(4)& 0.5920(4)&0.5909(4)&0.5939(3)&0.5901(4)&0.5582(5)&0.5670(5)&0.5919(4)& 0.5957(7)	&0.5930(3)&0.5755(5)&0.5922(4)\\
&& Phi-3-small & \cellcolor{red!40}0.6063(1)&	\cellcolor{red!40}0.6039(1)&	\cellcolor{red!40}0.6054(1)&	\cellcolor{red!40}0.6016(1)&	\cellcolor{red!40}0.6052(1)&\cellcolor{red!40}0.6061(1)&	\cellcolor{red!40}0.5688(1)&	\cellcolor{red!40}0.5893(1)&	\cellcolor{red!40}0.6048(1)&\cellcolor{red!40}0.6046(1)&\cellcolor{red!40}0.6006(1)&	\cellcolor{red!40}0.5964(1)&\cellcolor{red!40}0.6024(1)\\
&& Yi1.5 &  0.5911(7)&	0.5888(8)	&0.5919(5)	&0.5890(5)	&0.5906(5)	&0.5873(7)	&0.5589(4)	&0.5688(4)	&0.5912(5)	&0.5920(8)	&0.5905(6)	&0.5777(4)	&0.5906(5)\\
&& Internlm2.5 &  0.5914(5)&0.5890(7)	&0.5892(8)	&0.5888(6)	&0.5885(8)	&0.5852(8)	&0.5481(8)&	0.5657(7)&	0.5876(8)&	0.5898(5)	&0.5877(8)	&0.5696(8)	&0.5899(8)\\

\cmidrule(lr){2-16}
&
\multirow{8}{*}{LoRA} 
& Llama3.1 & 0.6083(2)&0.6062(3)&	0.6055(2)	&0.6068(2)	&0.6057(2)	&0.6053(3)	&0.5699(2)	&0.5900(3)&	\cellcolor{red!40}0.5976(1)	&0.6066(4)	&0.6055(3)&0.5990(5)&0.6017(3)\\
&& Qwen2.5 & \cellcolor{red!40}0.6090(1)&\cellcolor{red!40}0.6073(1)&	\cellcolor{red!40}0.6058(1)&	\cellcolor{red!40}0.6070(1)&\cellcolor{red!40}	0.6066(1)&	0.6070(2)&	0.5663(5)&	0.5927(2)&0.5932(3)		&\cellcolor{red!40}0.6090(1)&	0.6069(2)&\cellcolor{red!40}0.6058(1) & 0.6024(2)\\
&& Gemma2 & 0.6076(4)&0.6065(2)&	0.6050(4)&	0.6048(4)&	0.6049(3)&	\cellcolor{red!40}0.6072(1)&	0.5691(3)&	0.5881(4)&	0.5898(5)&	0.6075(2)&\cellcolor{red!40}	0.6071(1)&0.6051(2)&\cellcolor{red!40}0.6032(1)	\\
&& GLM4 & 0.6046(6)&0.6035(6)&	0.6045(6)&	0.6024(5)&	0.6020(6)&	0.6021(7)&	0.5639(7)&	0.5846(8)&	0.5917(4)&	0.6024(7)&	0.6023(7)&0.5994(4)& 0.5998(6)\\
&& Ministral & 0.6078(3)&0.6056(4)&0.6052(3)&0.6065(3)&0.6031(4)&0.6052(4)&0.5691(3)&0.5871(6)&	0.5943(2)&0.6070(3)&	0.6052(4)& 0.6036(3)&0.6012(4)\\
&& Phi-3-small & 0.6038(8)&0.6041(5)&0.6020(8)&0.6005(7)&0.6005(8)&0.6040(6)&\cellcolor{red!40}0.5723(1)&\cellcolor{red!40}0.5941(1)&0.5892(6)	&	0.6045(6)&0.6031(5)&0.5956(7) &0.5897(8)\\
&& Yi1.5 & 0.6065(5)&0.6033(7)	&0.6048(5)&	0.6023(6)&	0.6030(5)&	0.6046(5)&	0.5653(6)&	0.5878(5)&	0.5883(8)&	0.6051(5)&	0.6028(6)&0.5976(6) &0.6003(5)\\
&& Internlm2.5 & 0.6043(7)&0.6014(8)&0.6022(7)	&0.6001(8)&	0.6015(7)&	0.6019(8)&	0.5501(8)&	0.5855(7)&	0.5887(7)&	0.6022(8)&	0.6021(8)&0.5947(8)&0.5992(7)\\
\hline
\hline
\multirow{16}{*}{5\%} 
&
\multirow{8}{*}{Full} 
& Llama3.1 & 0.5978(2)&0.5944(3)&0.5937(3)&0.5941(3)&0.5936(3)&0.5925(3)&0.5562(3)&0.5686(4)&0.5957(2)&0.5937(3)&0.5799(3)&0.5956(3)&0.5940(3)\\
&& Qwen2.5 & 0.5973(3)&0.5949(2)&0.5938(2)&0.5965(2)&0.5943(2)&0.5947(2)&0.5580(2)&0.5719(2)&0.5945(3)&0.5967(2)&0.5836(2)&0.5973(2)&0.5958(2)\\
&& Gemma2 & 0.5898(8)&0.5889(6)&0.5910(6)&0.5899(6)&0.5862(8)&0.5891(6)&0.5512(6)&0.5610(8)&0.5903(6)&0.5888(7)&0.5777(5)&0.5893(7)&0.5889(8)\\
&& GLM4 &  0.5914(5)&0.5877(7)&0.5906(7)&0.5853(8)&0.5894(6)&0.5887(7)&0.5511(8)&0.5591(3)&0.5878(7)&0.5903(5)&0.5742(7)&0.5864(8)&0.5902(7)\\
&& Ministral &  0.5954(4) &0.5891(5)&0.5922(4)&0.5923(4)&0.5906(4)&0.5919(4)&0.5512(6)&0.5620(7)&0.5940(4)&0.5891(6)&0.5775(6)&0.5925(4)&0.5929(4)\\
&& Phi-3-small & \cellcolor{red!40}0.6063(1)&	\cellcolor{red!40}0.6034(1)&\cellcolor{red!40}	0.6036(1)&\cellcolor{red!40}	0.6014(1)&\cellcolor{red!40}	0.6047(1)&\cellcolor{red!40}	0.6050(1)&\cellcolor{red!40}	0.5603(1)&\cellcolor{red!40}	0.5865(1)&\cellcolor{red!40}	0.6045(1)&\cellcolor{red!40}	0.6032(1)	&\cellcolor{red!40}0.5955(1)	&\cellcolor{red!40}0.6001(1)&\cellcolor{red!40}	0.604(1)\\
&& Yi1.5 &  0.5911(7)&	0.5912(4)	&0.5916(5)&	0.5904(5)&	0.5895(5)&	0.5893(5)&	0.5529(4)&	0.5656(5)&	0.5936(5)&	0.5906(4)&	0.5781(4)&	0.5907(5)&	0.5922(5)\\
&& Internlm2.5 &  0.5914(5)&0.5875(8)&0.5885(8)	&0.5877(7)	&0.5887(7)&	0.5886(8)&	0.5424(5)&	0.5624(6)&	0.5874(8)&	0.5886(8)	&0.5732(8)&	0.5895(6)&	0.5917(6)\\
\cmidrule(lr){2-16}
&
\multirow{8}{*}{LoRA} 
& Llama3.1 & 0.6083(2)&0.6063(3)&0.6056(3)&0.6061(3)&0.6025(5)&\cellcolor{red!40}0.6091(1)&0.5638(3)&0.5881(3)&0.6068(2)&\cellcolor{red!40}0.6088(1)&\cellcolor{red!40}0.6078(1)&0.5984(4)&\cellcolor{red!40}0.6063(1)\\
&& Qwen2.5 & \cellcolor{red!40}0.6090(1)&0.6055(4)&0.6038(4)&0.6062(2)&0.6046(2)&0.6080(2)&0.5607(5)&0.5897(2)&0.6060(4)&0.6061(4)&0.6057(3)&\cellcolor{red!40}0.6007(1)&0.6045(4)\\
&& Gemma2 & 0.6076(4)&0.6074(2)&0.6058(2)&0.6050(4)&0.6044(3)&0.6014(8)&0.5639(2)&0.5730(8)&0.6068(2)&0.6072(3)&0.6044(4)&0.5996(3)&0.6060(3)\\
&& GLM4 & 0.6046(6)&0.6025(7)&0.6033(5)&0.6029(5)&0.6039(4)&0.6037(5)&0.5611(4)&0.5803(7)&0.6037(6)&0.6039(6)&0.6019(7)&0.5937(7)&0.6014(7)\\
&& Ministral & 0.6078(3)&\cellcolor{red!40}0.6077(1)&\cellcolor{red!40}0.6060(1)&\cellcolor{red!40}0.6076(1)&\cellcolor{red!40}0.6066(1)&0.6062(3)&0.5598(7)&0.5853(5)&\cellcolor{red!40}0.6100(1)&0.6079(2)&0.6074(2)&0.5970(6)&0.6062(2)\\
&& Phi-3-small &0.6038(8)&0.6027(6)&0.6027(6)&0.6007(8)&0.6021(6)&0.6031(7)&\cellcolor{red!40}0.5761(1)&\cellcolor{red!40}0.5923(1)&0.6029(8)&0.6031(7)&0.6027(5)&0.6001(2)&0.6034(5)\\
&& Yi1.5 & 0.6065(5)&0.6032(5)&0.6019(7)&0.6014(6)&0.6010(7)&0.6060(4)&0.5606(6)&0.5862(4)&0.6053(5)&0.6043(5)&0.6019(7)&0.5979(5)&0.6024(6)\\
&& Internlm2.5 & 0.5261(7)&0.6008(8)&0.6008(8)&0.6008(7)&0.6005(8)&0.6037(5)&0.5429(8)&0.5842(6)&0.6033(7)&0.6027(8)&0.6022(6)&0.5929(8)&0.6008(8)\\
\bottomrule
\end{tabular}}
\caption{The results of data filtering in the PubMedQA dataset.}
\label{table10}
\end{table*}

\renewcommand{\arraystretch}{0.90}
\begin{table*}[!th]
\centering
\scalebox{0.49}{
\begin{tabular}{cccccccc|cccc|cccc}
\toprule
\multirow{2}{*}{Sample} &
\multirow{2}{*}{Fine-tune} &
\multirow{2}{*}{Models} &
\multirow{2}{*}{GS} &
\multicolumn{4}{c}{Distribution-aware methods} &
\multicolumn{4}{c}{Duality-aware methods} &
\multicolumn{4}{c}{Both} \\
\cmidrule(lr){5-8} \cmidrule(lr){9-12} \cmidrule(lr){13-16} ratio
& & & & DQ & GraphCut & KCG & Random & AG & LMA & SF & LESS & MoDS & Cherry & Deita & CaR \\
\hline
\multirow{16}{*}{20\%} 
&
\multirow{8}{*}{Full} 
& Llama3.1 & 0.5621(6)	&0.5789(6)	&0.5732(6)	&0.5914(6)&	0.5932(5)&	0.5845(6)&	0.4612(7)&	0.6191(7)&	0.5743(4)&	0.5567(6)&	0.5912(6)&	0.6578(5)&	0.6123(6)  \\
&& Qwen2.5 &\cellcolor{red!40} 0.6885(1)&	\cellcolor{red!40}0.6845(1)&	\cellcolor{red!40}0.6871(1)&	0.6506(4)&	\cellcolor{red!40}0.7169(1)&	0.6867(2)&	0.6976(2)&	0.6712(3)&	\cellcolor{red!40}0.7134(1)&	0.6345(4)&	\cellcolor{red!40}0.6987(1)&	0.6789(3)&	0.6902(3)\\
&& Gemma2 & 0.5699(5)&	0.6301(5)&	0.6312(4)&	0.6578(3)&	0.4513(6)&	0.6245(5)	&0.6665(5)	&0.6513(4)&	0.5676(5)&	0.6176(5)&	0.6078(5)&	0.6123(6)	&0.6356(5)\\
&& GLM4 & 0.6497(3)	&0.6572(2)	&0.6545(3)	&0.6623(2)&0.6123(3)&\cellcolor{red!40}	0.6967(1)&\cellcolor{red!40}	0.7034(1)\cellcolor{red!40}	&\cellcolor{red!40}0.6936(1)	&0.5612(7)&	0.6812(2)&	0.6789(2)&	0.6876(2)&	\cellcolor{red!40}0.6978(1)\\
&& Ministral& 0.4033(7)&	0.4676(7)&	0.4523(7)	&0.4678(7)&	0.4312(7)&	0.5012(7)&	0.6145(6)	&0.5167(6)&	0.4832(6)&	0.3856(7)&	0.4712(7)&	0.5012(7)	&0.4912(7)\\
&& Phi-3-small& -& - &- &- &- &- &- &- &- & -& -& -&-\\
&& Yi1.5      & 0.6407(4)&	0.6534(4)	&0.5967(5)&	0.6237(5)&	0.6123(3)&	0.6645(4)	&0.6745(4)&	0.6724(5)&	0.6456(3)&\cellcolor{red!40}0.6813(1)&	0.6578(4)&	0.6623(4)&	0.6578(4)\\
&& Internlm2.5& 0.6685(2)&	0.6786(3)&	0.6856(2)&	\cellcolor{red!40}0.6890(1)&	0.6857(2)	&0.6710(3)&	0.6895(3)&	\cellcolor{red!40}0.6936(1)&	0.6878(2)&	0.6645(3)&	0.6687(3)&\cellcolor{red!40}	0.6923(1)&	0.6972(2)\\

\cmidrule(lr){2-16}
&
\multirow{8}{*}{LoRA} 
& Llama3.1 & 0.6855(6)&	0.6302(5)&	0.5734(6)&	0.6162(5)&	0.4534(6)&	0.5824(6)&	0.6156(6)&	0.6196(6)&	0.5743(4)&	0.5565(6)&	0.6087(5)&	0.6074(6)&	0.6123(6)\\
&& Qwen2.5 & 0.7298(2)&	0.6596(3)&	0.6541(3)&	0.6573(3)&	0.6847(2)&	0.6912(2)&	0.6867(3)&	0.6702(3)&	\cellcolor{red!40}0.7062(1)&	0.6802(2)&	\cellcolor{red!40}0.7002(1)&	0.6887(2)&	\cellcolor{red!40}0.6967(1)\\
&& Gemma2 & \cellcolor{red!40}0.7389(1)&	\cellcolor{red!40}0.6812(1)&\cellcolor{red!40}	0.6864(1)&	0.6614(2)&	\cellcolor{red!40}0.7145(1)&	\cellcolor{red!40}0.6935(1)&	0.6978(2)&	\cellcolor{red!40}0.6936(1)&	0.6867(2)&	0.6634(3)&	0.6789(2)&	\cellcolor{red!40}0.6921(1)&	0.6912(3)\\
&& GLM4 & 0.7254(3)&	0.6723(2)&	0.6834(2)&	\cellcolor{red!40}0.6887(1)&	0.6134(3)&	0.6674(3)&	\cellcolor{red!40}0.7123(1)&	0.6898(2)&	0.6445(3)&	\cellcolor{red!40}0.6834(1)&	0.6676(3)&	0.6786(3)&	0.6921(2)\\
&& Ministral & 0.6213(7)&	0.4245(7)&	0.4534(7)&	0.4621(7)&	0.4321(7)&	0.4937(7)&	0.4623(7)&	0.5223(7)&	0.4834(7)&	0.4021(7)&	0.4667(7)&	0.5012(7)&	0.4912(7)\\
&& Phi-3-small & -&-  & -& -& -&- &- &- &- &- &- &- &-\\
&& Yi1.5 & 0.6956(4)&	0.6612(4)&	0.6123(5)&	0.6509(4)&	0.5812(5)&	0.6624(4)&	0.6812(4)&	0.6656(4)&	0.5623(5)&	0.6176(5)&	0.5886(6)&	0.6678(4)&	0.6576(4)\\
&& Internlm2.5 & 0.6875(5)&	0.5821(6)&	0.6256(4)&	0.5923(6)&	0.5921(4)&	0.6278(5)&	0.6623(5)&	0.6536(5)&	0.5540(6)&	0.6345(4)&	0.6546(4)&	0.6612(5)&	0.6312(5)\\
\hline
\hline
\multirow{16}{*}{10\%} 
&
\multirow{8}{*}{Full} 
& Llama3.1 & 0.5621(6)&	0.5784(6)&	0.5720(6)&	0.5895(6)&	0.5792(5)&	0.5813(6)&	0.6101(6)	&0.6180(6)&	0.5631(4)&	0.5517(6)&	0.5877(6)&	0.6069(6)&	0.6108(6)\\
&& Qwen2.5 & \cellcolor{red!40}0.6885(1)&	\cellcolor{red!40}0.6800(1)&	\cellcolor{red!40}0.6850(1)&	0.6151(5)&	\cellcolor{red!40}0.7130(1)&	0.6860(2)&	0.6842(3)&	0.6671(3)&	\cellcolor{red!40}0.7048(1)&	0.6788(2)&	0.6778(2)&	0.6874(2)&	0.6892(3)\\
&& Gemma2 & 0.5699(5)&	0.6294(5)&	0.5952(5)&	0.6489(4)&	0.4485(6)&	0.6265(5)&	0.6610(5)&	0.6468(5)&	0.5606(5)&	0.6165(5)&	0.6076(5)&	0.6568(5)&	0.6301(5)\\
&& GLM4 & 0.6497(3)&	0.6564(3)&	0.6233(4)&	0.6600(2)&	0.6048(3)&	0.6650(3)&\cellcolor{red!40}	0.7030(1)&\cellcolor{red!40}	0.6924(1)&	55.39(6)&	0.6318(4)&	0.6664(3)&	\cellcolor{red!40}0.6899(1)	&\cellcolor{red!40}0.6938(1)\\
&& Ministral & 0.4033(7)	&0.4635(7)&	0.4485(7)&	0.4610(7)&	0.4211(7)&	0.4902(7)&	0.4571(7)&	0.5165(7)	&0.4752(7)&	0.3837(7)	&0.4652(7)&	0.4962(7)	&0.4894(7)\\
&& Phi-3-small & -& - &- &- &- &- &- &- &- &- &- &- &\\
&& Yi1.5 & 0.6407(4)	&0.6521(4)&	0.6532(3)&	0.6561(3)&	0.5841(4)&	0.6603(4)	&0.6728(4)&	0.6628(4)&	0.6415(3)&	0.6625(3)&	0.6532(4)	&0.6589(4)&	0.6546(4)\\
&& Internlm2.5 & 0.6685(2)&	0.6771(2)&	0.6829(2)&	\cellcolor{red!40}0.6867(1)	&0.6835(2)&	\cellcolor{red!40}0.6909(1)	&0.6963(2)&\cellcolor{red!40}	0.6924(1)&	0.6845(2)&\cellcolor{red!40}	0.6799(1)&\cellcolor{red!40}	0.6963(1)&	67.71(3)&	0.6917(2)\\

\cmidrule(lr){2-16}
&
\multirow{8}{*}{LoRA} 
& Llama3.1 & 0.6855(6)&	0.6835(5)&	0.6846(6)&	0.6824(5)&	0.6728(6)&	0.6693(6)&	0.6967(4)	&0.6803(5)&	0.6607(6)&	0.6461(6)&	0.6867(4)&	0.6817(5)&	0.6823(6)\\
&& Qwen2.5 & 0.7298(2)&	0.7234(2)&	0.7195(2)&	0.7134(2)&	0.7204(2)&	0.7212(2)&	0.7116(3)&	0.7124(3)&	0.6985(3)&	0.7052(3)&	0.7198(2)&	0.7124(3)&	0.7134(2)\\
&& Gemma2 & \cellcolor{red!40}0.7389(1)&	\cellcolor{red!40}0.7277(1)&	\cellcolor{red!40}0.7243(1)&	\cellcolor{red!40}0.7273(1)&	\cellcolor{red!40}0.7266(1)&	\cellcolor{red!40}0.7259(1)&	\cellcolor{red!40}0.7366(1)&	\cellcolor{red!40}0.7376(1)&	\cellcolor{red!40}0.7220(1)&	\cellcolor{red!40}0.7216(1)&	\cellcolor{red!40}0.7305(1)&	\cellcolor{red!40}0.7297(1)&	\cellcolor{red!40}0.7305(1)\\
&& GLM4 & 0.7254(3)&	0.7138(3)&	0.7063(3)&	0.7033(3)&	0.7123(3)&	0.7116(3)&	0.7180(2)&	0.7141(2)&	0.7013(2)&	0.7074(2)&	0.7063(3)&	0.7131(2)&	0.7012(3)\\
&& Ministral & 0.6213(7)&	0.6287(6)&	0.6323(7)&	0.6308(7)&	0.6225(7)&	0.6251(7)&	0.6472(7)&	0.6312(7)&	0.6209(7)&	0.6169(7)&	0.6286(7)&	0.6443(7)&	0.6383(7)\\
&& Phi-3-small &- & - & -& -& -& -&- &- &- &- &- &- &-\\
&& Yi1.5 & 0.6956(4)&	0.6006(7)&	0.6945(4)&	0.6945(4)&	0.6823(4)&	0.6881(4)&	0.6881(5)&	0.6878(4)&	0.6789(4)&	0.6867(4)&	0.6824(5)&	0.6864(4)&	0.6913(4)\\
&& Internlm2.5 & 0.6875(5)&	0.6597(4)&	0.6867(5)&	0.6782(6)&	0.6757(5)&	0.6718(5)&	0.6753(6)&	0.6654(6)&	0.6693(5)&	0.6703(5)&	0.6718(6)&	0.6734(6)&	0.6636(5)\\
\hline
\hline
\multirow{16}{*}{5\%} 
&
\multirow{8}{*}{Full} 
& Llama3.1 & 0.5621(6)	&0.5724(5)&	0.5692(6)&	0.6169(6)&	0.5977(6)&	0.5803(5)&	0.6027(6)&	0.6262(6)&	0.5453(5)	&0.6041(6)&	0.6180(6)&	0.5742(6)&	0.6141(6)\\
&& Qwen2.5 & \cellcolor{red!40}0.6885(1)&	0.6472(3)&	0.6518(3)&	0.6707(2)&	0.6472(3)	&0.6831(2)&	0.6329(5)&	0.6486(5)	&0.6262(4)&	0.6553(5)&	0.6578(4)&	\cellcolor{red!40}0.6985(1)	&0.6643(4)\\
&& Gemma2 & 0.5699(5)&	0.4136(6)&	0.6084(5)&	0.6489(3)	&0.6162(5)	&0.3271(6)	&0.6628(4)	&0.6635(3)	&0.5097(6)&	0.6596(4)	&0.6635(3)&	0.6073(5)&	0.6650(3)\\
&& GLM4 & 0.6497(3)&	0.6219(4)&	0.6593(2)&	0.6383(4)&	0.6500(2)&	0.6073(4)&\cellcolor{red!40}	0.7166(1)&\cellcolor{red!40}	0.7091(1)&	0.6333(3)&	0.6739(2)	&\cellcolor{red!40}0.6920(1)	&0.6201(4)&	\cellcolor{red!40}0.7052(1)\\
&& Ministral & 0.4033(7)	&0.4111(7)	&0.3271(7)&	0.2880(7)	&0.3335(7)	&0.2353(7)&	0.4898(7)&	0.4902(7)	&0.4905(7)&	0.4553(7)&	0.4749(7)&	0.2624(7)	&0.4913(7)\\
&& Phi-3-small & -& - &- &- & -& -& -& -& -& -&- &- &-\\
&& Yi1.5 & 0.6407(4)	&0.6521(2)&	0.6365(4)	&0.6265(5)&	0.6283(4)&	0.6621(3)&	0.6849(3)&	0.6546(4)&	0.6589(2)&	0.6642(3)&	0.6575(5)	&0.6508(3)&	0.6504(5)\\
&& Internlm2.5 & 0.6685(2)&\cellcolor{red!40}	0.6742(1)&\cellcolor{red!40}	0.6796(1)&\cellcolor{red!40}	0.6838(1)&	\cellcolor{red!40}0.6817(1)	&\cellcolor{red!40}0.6899(1)&	0.6909(2)&	0.6874(2)&	\cellcolor{red!40}0.6749(1)&	\cellcolor{red!40}0.6927(1)&	0.6899(2)&	0.6757(2)	&0.6828(2)\\
\cmidrule(lr){2-16}
&
\multirow{8}{*}{LoRA} 
& Llama3.1 &0.6855(6) & 0.6811(5)&	0.6789(5)	&0.6953(5)&	0.6853(4)&	0.6754(5)&	0.7056(4)&	0.7006(4)&	0.6547(6)&	0.6166(7)&	0.6842(5)&	0.6898(5)&	0.6906(4)\\
&& Qwen2.5 &0.7298(2) & 0.7184(2)&	0.7205(2)&	0.7102(3)&	0.7234(2)&	0.7188(2)&	0.7177(2)&	0.7170(3)&	0.7074(3)&	0.7220(2)&	0.7163(3)&	0.7141(3)&	0.7120(3)\\
&& Gemma2 &\cellcolor{red!40}0.7389(1) & \cellcolor{red!40}0.7333(1)	&\cellcolor{red!40}0.7345(1)&\cellcolor{red!40}	0.7330(1)&\cellcolor{red!40}	0.7316(1)&\cellcolor{red!40}	0.7305(1)	&\cellcolor{red!40}0.7387(1)&\cellcolor{red!40}	0.7319(1)&\cellcolor{red!40}	0.7220(1)&\cellcolor{red!40}	0.7255(1)&	\cellcolor{red!40}0.7312(1)&	\cellcolor{red!40}0.7245(1)&\cellcolor{red!40}	0.7322(1)\\
&& GLM4 &0.7254(3) & 0.7098(3)&	0.7123(3)&	0.7113(2)&	0.7074(3)&	0.7038(3)&	0.7173(3)&	0.7223(2)&	0.7063(2)&	0.6988(3)&	0.7145(2)&	0.7176(2)&	0.7205(2)\\
&& Ministral &0.6213(7) &0.6344(7)	&0.6478(7)	&0.6497(7)&	0.6469(7)&	0.6408(7)&	0.6433(7)	&0.6429(7)	&0.6194(7)	&0.6856(5)	&0.6654(7)	&0.6547(7)&	0.6468(7)\\
&& Phi-3-small & -& - &- &- &- &- &- &- &- &- &- &- &-\\
&& Yi1.5 &0.6956(4) & 0.6967(4)	&0.6934(4)	&0.6956(4)	&0.6534(6)	&0.6863(4)&	0.6913(5)&	0.6903(5)&	0.6789(4)&	0.6953(4)&	0.6881(4)&	0.6978(4)&	0.6857(5)\\
&& Internlm2.5 &0.6875(5) & 0.6711(6)&	0.6656(7)&	0.6775(6)&	0.6625(5)&	0.6686(6)&	0.6845(6)&	0.6735(6)&	0.6632(5)&	0.6728(6)&	0.6721(6)&	0.6713(6)&	0.6707(6)\\
\bottomrule
\end{tabular}}
\caption{The results of data filtering in the Alpaca dataset. Since the base model for Phi-3-small has not been publicly released, experiments were not conducted on this model using the Alpaca dataset. In the figure, "Ministral" refers to Mistral-7B-Instruct-v0.3.}
\label{table11}
\end{table*}

\renewcommand{\arraystretch}{0.90}
\begin{table*}[!th]
\centering
\scalebox{0.50}{
\begin{tabular}{cccccccc|cccc|cccc}
\toprule
\multirow{2}{*}{Sample} &
\multirow{2}{*}{FT} &
\multirow{2}{*}{Models} &
\multirow{2}{*}{Baseline} &
\multicolumn{4}{c}{Distribution-aware methods} &
\multicolumn{4}{c}{Quality-aware methods} &
\multicolumn{4}{c}{Both} \\
\cmidrule(lr){5-8} \cmidrule(lr){9-12} \cmidrule(lr){13-16} ratio
& & & & DQ & GraphCut & KCG & Random & AG & LMA & SF & LESS & MoDS & Cherry & Deita & CaR \\
\hline
\multirow{16}{*}{20\%} 
&
\multirow{8}{*}{Full} 
& Llama3.1 & \cellcolor{red!40}0.6586(1) &	0.6502(3)&	0.6514(3)&	0.6596(2)&	0.6582(2)&	0.6438(4)&	0.6463(2)&	0.6141(4)&\cellcolor{red!40}	0.6465(1)&	\cellcolor{red!40}0.6576(1)&	0.6533(3)&	0.6544(3)&\cellcolor{red!40}	0.6652(1)  \\
&& Qwen2.5 &0.6581(3)&	0.6547(2)&	0.6504(4)&	0.6527(3)&	0.6491(4)&	\cellcolor{red!40}0.6643(1)&	0.6366(4)&	0.6273(3)&	0.6267(4)&	0.6473(4)&\cellcolor{red!40}	0.6632(1)&	0.6563(2)&	0.6484(3)\\
&& Gemma2 & 0.6428(6)& 	0.6402(6)&	0.6475(5)&	0.6502(4)&	0.6485(6)&	0.6383(6)&	0.6294(6)&	0.6103(6)&	0.6245(5)&	0.6354(6)&	0.6482(6)&	0.6482(6)&	0.6465(4)\\
&& GLM4 & 0.5927(8)&	0.6022(8)&	0.5875(8)&	0.6045(8)&	0.6105(8)&	0.6017(8)&	0.6152(8)&	0.5788(8)&	0.5873(8)&	0.5891(8)&	0.6158(8)&	0.6182(8)&	0.5973(8)\\
&& Ministral& 0.6191(7)&	0.6173(7)&	0.6186(7)&	0.6184(7)&	0.6147(7)&	0.6166(7)&	0.6215(7)&	0.5984(7)&	0.6043(7)&	0.6054(7)&	0.6217(7)&	0.6308(7)&	0.6032(7)\\
&& Phi-3-small& 0.6514(5) &	0.6476(4)&	0.6533(2)&	0.6498(5)&	0.6456(5)&	0.6422(5)&	0.6403(3)&	0.6265(2)&	0.6177(6)&	0.6483(3)&	0.6504(4)&	0.6506(5)&	0.6432(5)\\
&& Yi1.5      & 0.6550(4)&	0.6453(5)&	0.6476(5)&	0.6477(6)&	0.6534(3)&	0.6509(3)&	0.6312(5)&	0.6103(5)&	0.6326(3)&	0.6424(5)&	0.6496(5)&\cellcolor{red!40}	0.6652(1)&	0.6426(6)\\
&& Internlm2.5& \cellcolor{red!40}0.6586(1)&	\cellcolor{red!40}0.6603(1)&\cellcolor{red!40}	0.6558(1)&\cellcolor{red!40}	0.6623(1)&\cellcolor{red!40}	0.6656(1)&	0.6530(2)&\cellcolor{red!40}	0.6574(1)&\cellcolor{red!40}	0.6373(1)&	0.6373(2)&	0.6514(2)&	0.6577(2)&	0.6529(4)&\cellcolor{red!40}	0.6652(1)\\

\cmidrule(lr){2-16}
&
\multirow{8}{*}{LoRA} 
& Llama3.1 & 0.6624(5)&\cellcolor{red!40}	0.6592(1)&	0.6493(5)	&0.6515(3)&	0.6532(2)&	0.6464(3)&	0.6406(3)&	0.6257(2)&	0.6264(4)&\cellcolor{red!40}	0.6573(1)&	0.6577(3)&	0.6577(2)&\cellcolor{red!40}	0.6593(1)\\
&& Qwen2.5 & 0.6633(4)&	0.6434(5)&	0.6506(4)&	0.6485(5)&	0.6473(5)&	0.6422(5)&	0.6315(4)&	0.6143(5)&	0.6254(5)&	0.6428(5)&	0.6522(4)&	0.6501(5)&	0.6427(5)\\
&& Gemma2 & 0.6669(3)&	0.6522(2)&	0.6548(2)&	0.6506(4)&	0.6509(3)&	0.6503(2)&	0.6315(4)&	0.6177(4)&	0.6385(2)&	0.6458(4)&	0.6583(2)&	0.6547(4)&	0.6473(4)\\
&& GLM4 & 0.6685(2)&	0.6495(3)&0.6527(3)&	0.6545(2)&	0.6482(4)&	0.6446(4)&\cellcolor{red!40}	0.6584(1)&\cellcolor{red!40}	0.6353(1)&\cellcolor{red!40}	0.6485(1)&	0.6467(3)&	0.6495(5)&\cellcolor{red!40}	0.6613(1)&	0.6558(2)\\
&& Ministral & \cellcolor{red!40}0.6707(1)	&0.6455(4)&	\cellcolor{red!40}	0.6572(1)&	\cellcolor{red!40}0.6561(1)&	\cellcolor{red!40}0.6545(1)&\cellcolor{red!40}	0.6578(1)&	0.6473(2)&	0.6185(3)&	0.6333(3)&	0.6496(2)&\cellcolor{red!40}	0.6624(1)&	0.6566(3)&	0.6538(3)\\
&& Phi-3-small & 0.6544(8)&	0.6034(8)&	0.6026(8)&	0.6234(7)&	0.6093(8)&	0.6154(7)&	0.6236(7)&	0.5407(8)&	0.5994(7)&	0.6123(7)&	0.6237(7)&	0.6126(8)&	0.5904(8)\\
&& Yi1.5 & 0.6573(6)&	0.6406(6)&	0.6484(6)&	0.6065(8)&	0.6453(6)&	0.6364(6)&	0.6155(8)&	0.6104(6)&	0.5877(8)&	0.6355(6)&	0.6128(8)&	0.6493(6)&	0.6405(6)\\
&& Internlm2.5 & 0.6565(7)&	0.6245(7)&	0.6233(7)&	0.6486(6)&	0.6123(7)&	0.5904(8)&	0.6295(6)&	0.5982(7)&	0.6185(6)&	0.5937(8)&	0.6475(6)&	0.6303(7)&	0.5956(7)\\
\hline
\hline
\multirow{16}{*}{10\%} 
&
\multirow{8}{*}{Full} 
& Llama3.1 & \cellcolor{red!40}0.6586(1) &	0.6455(3)&	0.6507(3)&	0.6503(2)&	\cellcolor{red!40}0.6507(1)&	0.6376(5)&	\cellcolor{red!40}0.6486(1)&	0.6054(4)&	0.6226(5)&	\cellcolor{red!40}0.6486(1)&	0.6495(4)&	0.6537(3)&	0.6398(5)\\
&& Qwen2.5 & 0.6581(3)&	0.6494(2)&	0.6497(4)&	0.6493(3)&	\cellcolor{red!40}0.6507(1)&	0.6495(2)&	0.6378(2)&	0.6102(3)&	0.6234(4)&	0.6445(3)&\cellcolor{red!40}	0.6551(1)&\cellcolor{red!40}	0.6563(1)&	0.6427(4)\\
&& Gemma2 & 0.6428(6)& 	0.6423(5)&	0.6484(5)&	0.6485(4)&	0.6416(6)&	0.6434(3)&	0.6206(7)&	0.6045(5)&	0.6106(6)&	0.6308(6)&	0.6446(6)&	0.6483(5)&	0.6445(3)\\
&& GLM4 & 0.5927(8)&	0.6000(8)&	0.6013(8)&	0.6014(8)&	0.6072(7)&	0.5882(8)&	0.6264(6)&	0.5373(8)&	0.5704(8)&	0.5791(8)&	0.6104(8)&	0.6083(8)&	0.5865(8)\\
&& Ministral & 0.6191(7)	&0.6142(7)	&0.6175(7)&	0.6176(7)&	0.6054(8)&	0.6043(7)&	0.6122(8)&	0.5685(7)&	0.5924(7)&	0.5973(7)&	0.6196(7)&	0.6237(7)&	0.5913(7)\\
&& Phi-3-small & 0.6514(5)& \cellcolor{red!40}	0.6573(1)&	\cellcolor{red!40}0.6517(1)&	0.6464(5)&	0.6438(5)&	0.6349(6)&	0.6293(4)&\cellcolor{red!40}	0.6214(1)&	\cellcolor{red!40}0.6395(1)&	0.6447(3)&	0.6503(3)&	0.6545(2)&	0.6528(2)\\
&& Yi1.5 & 0.6550(4)&	0.6395(6)&	0.6454(6)&	0.6457(6)&	0.6448(4)&	0.6379(4)&	0.6262(5)&	0.6025(6)&	0.6266(3)&	0.6417(5)&	0.6464(5)&	0.6473(6)&	0.6384(6)\\
&& Internlm2.5 & \cellcolor{red!40}0.6586(1)&	0.6455(3)&\cellcolor{red!40}	0.6517(1)&\cellcolor{red!40}	0.6514(1)&	0.6476(3)&	\cellcolor{red!40}0.651(1)&	0.6358(3)&	0.6139(2)&	0.6270(2)&	0.6454(2)&	0.6512(2)&	0.6515(4)&	\cellcolor{red!40}0.6546(1)\\

\cmidrule(lr){2-16}
&
\multirow{8}{*}{LoRA} 
& Llama3.1 & 0.6624(5)& \cellcolor{red!40} 0.6614(1)&	0.6563(4)&	0.6536(2)	&0.6593(2)&	0.6556(3)&	0.6559(3)	&0.6072(6)&	0.6351(5)&\cellcolor{red!40}	0.6595(1)&	0.6594(3)&	0.6573(2)&	\cellcolor{red!40}0.6676(1)\\
&& Qwen2.5 &0.6635(4) &  0.6513(4)	&0.6515(5)&	0.6294(7)&	0.6573(3)&	0.6467(7)&	0.6508(4)&	0.6080(4)&	0.6372(4)&	0.6505(5)&	0.6516(6)&	0.6477(6)&	0.6622(5)\\
&& Gemma2 &0.6669(3) &  0.6547(3)&	0.6648(2)&	0.6455(4)&	0.6559(4)&	0.6550(3)&	0.6344(5)&	0.6235(2)&	0.6397(2)&	0.6498(6)&	0.6629(2)&	0.6513(4)&	0.6665(3)\\
&& GLM4 &0.6685(2) &  0.6585(2)&0.6627(3)&	0.6473(3)&	0.6534(5)&	0.6562(2)&	0.6575(2)&\cellcolor{red!40}	0.6267(1)&	\cellcolor{red!40}0.6425(1)&	0.6543(2)&	0.6586(4)&	\cellcolor{red!40}0.6647(1)&	0.6663(3)\\
&& Ministral &\cellcolor{red!40}0.6707(1) &  0.6035(8)	&\cellcolor{red!40}	0.6673(1)&\cellcolor{red!40}	0.6566(1)&\cellcolor{red!40}	0.6604(1)&	\cellcolor{red!40}0.6645(1)&	\cellcolor{red!40}0.6632(1)&	0.6084(4)&	0.6295(7)&	0.6547(2)&	\cellcolor{red!40}0.6645(1)&	0.6566(3)&	\cellcolor{red!40}0.6676(1)\\
&& Phi-3-small &0.6544(8) &  0.6073(7)&	0.6266(7)&	0.6237(8)&	0.6375(8)&	0.3623(8)&	0.6256(7)&	0.6153(3)&	0.6196(8)&	0.6105(8)&	0.6423(8)&	0.6436(7)	&0.6482(8)\\
&& Yi1.5 &0.6573(6) &  0.6493(5)&	0.6512(5)	&0.6375(6)&	0.6496(7)	&0.6515(5)	&0.6133(8)	&0.6027(8)&	0.6338(6)	&0.6522(4)	&0.6524(5)&	0.6526(5)&	0.6604(6)\\
&& Internlm2.5 &0.6565(7) & 0.6493(6)&	0.6495(6)&	0.6444(5)&	0.6522(6)&	0.6496(6)&	0.6334(6)&	0.6036(7)&	0.6394(2)&	0.6456(7)&	0.6463(7)&	0.6421(8)	&0.6524(7)\\
\hline
\hline
\multirow{16}{*}{5\%} 
&
\multirow{8}{*}{Full} 
& Llama3.1 & \cellcolor{red!40}0.6586(1)& \cellcolor{red!40}	0.6463(1)&	0.6464(3)&	0.6332(5)&	0.6455(3)&	0.6436(4)&	0.6254(2)&	0.5875(5)&	0.6376(3)&	0.6433(5)&	0.6536(4)&	0.5922(5)&	0.6565(2)\\
&& Qwen2.5 & 0.6581(3)&	0.6403(4)&	\cellcolor{red!40}0.6536(1)&	0.6385(3)&	\cellcolor{red!40}0.6483(1)&	0.6494(2)&\cellcolor{red!40}	0.6293(1)&	0.6052(2)&	0.6345(4)&	0.6486(3)&	\cellcolor{red!40}0.6597(1)&	0.6024(3)&	0.6545(4)\\
&& Gemma2 & 0.6428(6)&	0.6375(6)&	0.6436(6)&	0.6357(4)&	0.6423(5)&	0.6476(3)&	0.6041(5)&	0.5843(6)&	0.6242(6)&	0.6334(6)&	0.6553(3)&	0.5842(6)&	0.6534(5)\\
&& GLM4 & 0.5927(8)&	0.6013(8)&	0.5982(8)&	0.5834(8)&	0.6055(8)&	0.6086(7)&	0.0493(7)&	0.5245(8)&	0.5866(8)&	0.6124(8)&	0.6145(8)&	0.5597(8)&	0.6124(8)\\
&& Ministral & 0.6191(7)&	0.6033(7)&	0.6075(7)&	0.6036(7)&	0.6104(7)&	0.6086(7)&	0.5873(8)&	0.5435(7)&	0.5936(7)&	0.6187(7)&	0.6273(7)&	0.5675(7)&	0.6146(7)\\
&& Phi-3-small & 0.6514(5)&	0.6454(2)&	0.6453(4)&\cellcolor{red!40}	0.6495(1)&	0.6446(4)&	0.6337(6)&	0.5924(6)&	\cellcolor{red!40}0.6143(1)&	\cellcolor{red!40}0.6445(1)&	0.6466(4)&	0.6513(6)&	\cellcolor{red!40}0.6106(1)	&\cellcolor{red!40}0.6624(1)\\
&& Yi1.5 & 0.6550(4)&	0.6392(5)&	0.6446(5)&	0.6314(6)&	0.6346(6)&	0.6424(5)&	0.6122(4)&	0.5936(4)&	0.6294(5)&	\cellcolor{red!40}0.6538(1)&	0.6569(2)&	0.5934(4)&	0.6516(6)\\
&& Internlm2.5 & \cellcolor{red!40} 0.6586(1)&	0.6433(3)&	\cellcolor{red!40}0.6536(1)&	0.6407(2)&\cellcolor{red!40}	0.6483(1)&	\cellcolor{red!40}0.6507(1)&	0.6214(3)&	0.6033(3)&	0.6396(2)&	0.6497(2)&	0.6538(4)&	0.6064(2)&	0.6557(3)\\
\cmidrule(lr){2-16}
&
\multirow{8}{*}{LoRA} 
& Llama3.1 &0.6624(5) &  0.6558(3)&	0.6475(5)&	0.6454(7)&	0.6533(3)&	0.6605(3)&	\cellcolor{red!40}0.6476(1)	&0.6103(4)	&0.6164(7)&	0.6575(2)&	0.6576(2)&	\cellcolor{red!40}0.6643(1)&	0.6675(2)\\
&& Qwen2.5 &0.6635(4) &0.6545(4)	&0.6524(3)&	0.6516(5)	&0.6532(3)&	0.6515(7)	&0.6285(5)&	0.6102(4)	&0.6195(5)&	0.6466(6)&	0.6507(6)	&0.6565(4)&	0.6587(5)\\
&& Gemma2 &0.6669(3) &  0.6526(6)	&0.6425(6)&	0.6558(2)&	0.6524(5)&\cellcolor{red!40}	0.6633(1)&	0.6426(2)&	0.6147(2)	&0.6284(2)&\cellcolor{red!40}	0.6585(1)	&0.6546(3)&	0.6583(3)&\cellcolor{red!40}	0.6696(1)\\
&& GLM4 &0.6685(2) &  \cellcolor{red!40}0.6642(1)	&0.6554(2)	&\cellcolor{red!40}0.6624(1)&	0.6597(2)	&0.6606(3)	&0.6357(3)&	\cellcolor{red!40}0.6336(1)&	\cellcolor{red!40}0.6293(1)&	0.6578(2)&\cellcolor{red!40}	0.6648(1)	&0.6627(2)	&0.6614(4)\\
&& Ministral &\cellcolor{red!40}0.6707(1) &  0.6595(2)&\cellcolor{red!40}	0.6575(1)&	0.6547(3)&	\cellcolor{red!40}0.6612(1)&	0.6615(2)&	0.6307(4)&	0.6035(5)&	0.6207(3)&	0.6534(4)&	0.6547(3)	&0.6497(5)&	0.6635(3)\\
&& Phi-3-small &0.6544(8) & 0.6254(8)	&0.6353(7)&	0.6395(8)&	0.6183(8)	&0.6245(8)&	0.6192(7)&	0.5704(8)&	0.5915(8)&	0.5836(8)&	0.6447(8)&	0.6448(8)&	0.6444(8)\\
&& Yi1.5 &0.6573(6) &  0.6546(4)&	0.6507(4)&	0.6518(5)	&0.6506(7)&	0.6531(6)&	0.6164(8)	&0.6118(3)	&0.6153(6)&	0.6497(5)	&0.6524(5)	&0.6490(5)&	0.6584(5)\\
&& Internlm2.5 &0.6565(7) & 0.6495(7)	&0.6333(8)	&0.6542(3)&	0.6514(6)&	0.6585(5)&	0.6226(6)&	0.5976(7)&	0.6204(3)	&0.6463(6)&	0.6502(6)&	0.6471(7)&	0.6475(7)\\
\bottomrule
\end{tabular}}
\caption{The results of data filtering in the SQuAD dataset.}
\label{table12}
\end{table*}

\section{Additional Results for Ablation Study}
\label{appdixE}

Tables~\ref{table13} and~\ref{table14}  present the results of ablation studies on different datasets.

\end{document}